\documentclass[journal]{IEEEtran}
\usepackage{epsfig,amsmath,amssymb,epsf,cite,scalefnt,subfig,array} %caption2,
\usepackage{cases}
\usepackage{graphicx,pifont,epstopdf,grffile}
\usepackage{multirow}

\def\bz{{\mathbf{z}}}

\def\bA{{\mathbf{A}}}  \def\bC{{\mathbf{C}}} \def\bD{{\mathbf{D}}} 
   \def\bI{{\mathbf{I}}} 
    
  \def\bR{{\mathbf{R}}}  
   \def\bX{{\mathbf{X}}}

 \def\ibb{{\pmb{b}}} \def\ibc{{\pmb{c}}} \def\ibd{{\pmb{d}}}

\def\ibu{{\pmb{u}}}  \def\ibw{{\pmb{w}}} \def\ibx{{\pmb{x}}} \def\iby{{\pmb{y}}}
\def\ibz{{\pmb{z}}}

\def\argmin{\mathop{\mathrm{argmin}}}

\renewcommand{\baselinestretch}{1.0}

     \def\d4{\!\!\!\!}

\def\bal{\boldsymbol{\alpha}} \def\blam{\boldsymbol{\lambda}}

\def\bpsi{\mathbf{\psi}}
\def\bupsi{\boldsymbol{\upsilon}}
  \def\R{{\mathbb{R}}}

\newtheorem{proposition}{Proposition}
\newtheorem{theorem}{Theorem}
\newtheorem{lemma}{Lemma}

\newcommand{\bef}{\begin{figure}}
\newcommand{\eef}{\end{figure}}
\newcommand{\beq}{\begin{eqnarray}}
\newcommand{\eeq}{\end{eqnarray}}

\newcommand{\qed}{\nobreak \ifvmode \relax \else
\ifdim\lastskip<1.5em \hskip-\lastskip \hskip1.5em plus0em
minus0.5em \fi \nobreak \vrule height0.5em width0.5em
depth0.25em\fi}

\ifCLASSINFOpdf
\else
\fi
\begin{document}
\title{$l_0$-norm Based Algorithm for Training Fault Tolerant RBF Networks and Selecting Centers}

\author{Hao~Wang,
        Chi-Sing~Leung,~\IEEEmembership{Member,~IEEE,}
        Hing Cheung So,~\IEEEmembership{Fellow,~IEEE,}
        Ruibin~Feng,
        and~Zifa~Han
\thanks{Hao~Wang, Chi-Sing~Leung, Hing Cheung So, Ruibin~Feng, and~Zifa~Han are with the Department of Electronic Engineering, City University of Hong Kong Kowloon Tong, Kowloon, Hong Kong.}}


\markboth{IEEE TRANSACTIONS ON NEURAL NETWORKS and LEARNING SYSTEMS, Vol.~, No.~, Month~2019}%
{Shell \MakeLowercase{\textit{et al.}}: $l_0$-norm Based Algorithm for Training Fault Tolerant RBF Networks and Selecting Centers}



\maketitle

\begin{abstract}
The aim of this paper is to train an RBF neural network and select centers under concurrent faults.
It is well known that fault tolerance is a very attractive property for neural networks.
And center selection is an important procedure during the training process of an RBF neural network.
In this paper, we devise two novel algorithms to address these two issues simultaneously.
Both of them are based on the ADMM framework.
%For both two methods, we first define a fault tolerant objective function. %based on all input vector in training set.
In the first method, the minimax concave penalty (MCP) function is introduced to select centers.
In the second method,  an $l_0$-norm term is directly used, and the hard threshold (HT) is utilized to address the $l_0$-norm term. Under several mild conditions, we can prove that both methods can globally converge to a unique limit point.
Simulation results show that, under concurrent fault, the proposed algorithms are superior to many existing methods.
\end{abstract}


% \fi
\IEEEpeerreviewmaketitle

\section{Introduction} \label{section1}

%\IEEEPARstart{T}{he}
Radial basis function (RBF) neural network is a common algorithm and is widely used in many applications~\cite{RBF2014,RBF2013,eickhoff2007robustness}.
Its training process usually includes two stages.
In the first phase, the RBF centers are determined.
And in the second phase, the corresponding weights of these RBF centers are estimated.
These RBF centers are usually selected from the training set. For instance, we can use all input vectors from the training samples as RBF centers~\cite{Poggio:1990}, or randomly choose a subset from the training set~\cite{Haykin:1998}. However, the first method may result in a complex network structure and the problem of overfitting.
The second method cannot guarantee that the constructed RBF network covers the input space well.

To overcome the shortcomings of the above two methods, many other RBF center selection approaches have been proposed. Among them, clustering algorithm~\cite{Chen:1995}, orthogonal least squares (OLS) approach~\cite{Chen:1991,Gomm:2000}, and support vector regression~\cite{Vapnik:1995,smola1997support} are the most representative methods.
None of these algorithms involve situations where network faults have occurred. However, during the training process of neural networks, the network faults are almost inevitable.
For example, when we calculate the centers' weights, the round-off errors will be introduced which can be seen as a kind of multiplicative weight fault~\cite{Stevenson1990,BurrJ1995,han2015}. When the connection between two neurons is damaged, signals cannot transform between them which may result in the open weight fault~\cite{rbf_fault3,nawrocki2011artificial}.

Over the past two decades, several fault tolerant neural networks have been proposed.
Most of them only consider one kind of network fault~\cite{paper:bernier2000b,Leung2008,Conti2000125,fnn_fault2}.
However, the paper \cite{paper:leung2012} first describes a situation when the multiplicative weight fault and open weight fault occur in a neural network concurrently.
Due to the modification of its objective function, the solution of this method is biased.
To handle this issue, a new approach based on OLS and regularization term is proposed in~\cite{leung2017regularizer}.
The performance of this algorithm is better than most existing methods.
But the computational cost is very expensive, since the OLS approach is used in this method.
And this method can only carry out the center selection steps before the training process.
In order to further improve the performance of an RBF network and perform center selection and training at the same time, a fault tolerant RBF center selection method based on $l_1$-norm is proposed in our previous work~\cite{wang2017admm}.

In this paper, we further develop our previous work by replacing the $l_1$-norm regularization with the $l_0$-norm term. And then we propose two methods to solve this problem.
In the first one, we further modify the objective by introducing the minimax concave penalty (MCP) function to substitute the $l_0$-norm term. After that, the problem is solved by alternating direction method of multipliers (ADMM) framework. In the second method, the ADMM framework and hard threshold (HT) are utilized to solve the problem. The main contribution of this paper is:
(i). Two novel fault tolerant RBF center selection algorithms are developed.
(ii). We theoretically prove that the proposed methods can globally converge to a unique limit point.
(iii). The performance improvements of the proposed methods are very significant.

The rest of paper is organized as follows. The background of RBF neural network under concurrent faults and the ADMM framework are described in Section~\ref{section2}. In Section~\ref{section3}, the proposed two approaches are developed. In Section~\ref{section4}, we prove that both our proposed methods can globally converge to a unique limit point under mild conditions. Numerical results for evaluation and comparison of different algorithms are provided in Section~\ref{section5}. Finally, the conclusions are drawn in Section~\ref{section6}.

\section{Background} \label{section2}
\subsection{Notation}
We use a lower-case or upper-case letter to represent a scalar while vectors and matrices are denoted by bold lower-case and upper-case letters, respectively. The transpose operator is denoted as $(\centerdot)^ \mathrm{T}$, and $\bI$ represents the identity matrix with appropriate dimensions. Other mathematical symbols are defined in their first appearance.

\subsection{RBF networks under concurrent fault situation}
In this paper, the training set is expressed as
\beq
\mathcal{D}=\left\{\left(\ibx_i,y_i\right):\ibx_i\in \R^{K_1}, y_i \in \R,
i=1,2,...,N \right\}\,, \label{trainning_set}
\eeq
where $\ibx_i$ is the input of the $i$-th sample with dimension $K_1$, and $y_i$ is the corresponding output. Similarly, the test set can be denoted as
\beq
\mathcal{D}' = \left\{ \left(\ibx'_{i'},y'_{i'}\right):\ibx'_{i'}
\in \R^{K_1}, y'_{i'} \in \R, i'=1,2,...,N' \right\}.
\label{test_set}
\eeq

Generally speaking, the RBF approach is used to handle regression problems. The input-output relationship of data in $\mathcal{D}$ is approximated by the sums of $M$ radial basis functions, i.e.,
\beq
f(\ibx) = \sum^M_{j=1} w_j \exp\left(\frac{-\left\|\ibx-\ibc_j\right\|_2^2}{s}\right) \, ,
\eeq
where $\exp\left(-\left\|\ibx-\ibc_j\right\|_2^2/s\right)$ is the $j$-th radial basis function, $w_j$ denotes its weight,
the vectors $\ibc_j$'s are the RBF centers, $s$ is a parameter which can be used to control the radial basis function width, and $M$ denotes the number of RBF centers.
Usually, the centers are selected from the input data $\{\ibx_1, \dots,\ibx_N\}$.
If we directly use all inputs as centers, it may result in some ill-conditioned solutions. Therefore, center selection is a key step in RBF neural network.
%OLS \cite{} and SVR \cite{} are two typical algorithms which can be used to handle this issue. Besides, in our previous work, we use a $l_1$-norm regulation term to do center selection \cite{}.

For a faulty-free network, the training set error can be expressed as
\beq
{\cal E}_{train} &=& \frac{1}{N}\sum\limits_{i=1}^N\left(y_i-f(\ibx_i)\right)^2\nonumber \\
&=& \frac{1}{N}\sum\limits_{i=1}^N \left(y_i - \sum^M_{j=1} w_j \exp\left(\frac{-\left\|\ibx_i-\ibc_j\right\|_2^2}{s}\right)\right)^2 \nonumber \\
&=&\frac{1}{N}\left\| \iby- \bA \ibw\right\|_2^2,
\eeq
where $\ibw=[w_1,\cdots,w_M]^\mathrm{T}$, $\iby=[y_1,\cdots,y_N]^\mathrm{T}$, and $\bA$ is a $N\times M$ matrix. Let $a_j(\ibx_i)$ denotes the $(i,j)$ entry of $\bA$,
\beq
a_j(\ibx_i)=[\bA]_{i,j}=\exp\left(-\frac{\left\|\ibx_i-\ibc_j\right\|_2^2}{s}\right).
\eeq

However, in the implementation of an RBF neural network, weight failures may happen. Multiplicative weight noise and open weight fault are two common faults in the RBF neural network \cite{Stevenson1990,BurrJ1995,han2015,paper:bernier2000b,Conti2000125,fnn_fault2, paper:Burr91digitalneural, Bernier2000c}.
When they are concurrent~\cite{paper:leung2012,leung2017regularizer}, the weights can be modeled as
\beq\label{eq:weight_con}
\tilde{w}_{j}=\left( w_j + b_j w_j \right)\beta_j,
\eeq
where $j=1,\cdots,M$, $\beta_j$ denotes the open fault of the $j$th weight. When the connection is opened, $\beta_j=0$, otherwise, $\beta_j=1$.
The term $b_j w_j$ in (\ref{eq:weight_con}) is the multiplicative noise in $j$th weight. The magnitude of the noise is proportional to that of the weight.
We assume that the $b_j$'s are independent and identically distributed (i.i.d.) zero-mean random variables with variance $\sigma_b^2$.
With this assumption, the statistics of  $b_j$'s are summarized as
\begin{subequations}
\beq
\langle b_j \rangle = 0, ~\langle b^2_j \rangle =\sigma_b^2, \\
\langle b_j b_{j'} \rangle = 0, \, \, \, \, \forall\, j \neq j',
\eeq
\label{statistics_b}
\end{subequations}
where $\langle \cdot \rangle$ is an expectation operator.
Furthermore,  we assume that $\beta_j$'s are i.i.d. binary random variables.
The probability mass function of $\beta_j$ is given by
\begin{numcases}{\mbox{Prob}(\beta_j)=}
P_{\beta}, & for $\beta_j= 0$,\\
1-P_{\beta}, & for $\beta_j=1$.
\end{numcases}
Thus, the statistics of $\beta_i$'s are
\begin{subequations}
\beq
&\langle \beta_j \rangle =\langle \beta^2_j \rangle =1-P_{\beta}, \\
&\langle \beta_j \beta_{j'} \rangle =(1-P_{\beta})^2, \, \, \, \, \forall\, j \neq j'.
\eeq
\label{statistics_beta}
\end{subequations}

Given a particular fault pattern of $b_j$ and $\beta_j$, the training set error can be expressed as
\beq \label{eq:training}
\tilde{\cal{E}}_{train} &=& \frac{1}{N} \left\| \iby- \bA \tilde{\ibw}\right\|_2^2 \nonumber\\
& = & \frac{1}{N} \sum\limits_{i=1}^N \left[  y_i^2 - 2y_i\sum\limits_{j=1}^M\beta_jw_j a_j(\ibx_i) \right.\nonumber\\
&& + \sum\limits_{j=1}^M\sum\limits_{j'=j}^M \beta_j\beta_{j'}w_j w_{j'}(1+b_j b_{j'})a_j(\ibx_i)a_{j'}(\ibx_i) \nonumber \\
&& + \sum\limits_{j=1}^M\sum\limits_{j'=1}^M (b_j+b_{j'})\beta_j\beta_{j'}w_j w_{j'}a_j(\ibx_i)a_{j'}(\ibx_i) \nonumber \\
&& \left.- 2y_i\sum\limits_{j=1}^Mb_j\beta_j w_j a_j(\ibx_i)\right].
\eeq
From \eqref{statistics_b} and \eqref{statistics_beta}, the average training set error~\cite{leung2017regularizer} over all possible failures is given by
\beq \label{eq:training2}
\overline{\cal E}_{train} = \frac{P_{\beta}}{N}\sum\limits_{i=1}^N y_i^2 +
\frac{1-P_{\beta}}{N} \left\| \iby - \bA \ibw \right\|_2^2 \nonumber\\
+ \frac{1-P_{\beta}}{N}\ibw^\mathrm{T}\left[(P_{\beta}+\sigma_b^2)\mbox{diag} \left(\bA^\mathrm{T} \bA\right)-P_{\beta}\bA^\mathrm{T} \bA\right]\ibw.
\eeq
%Note that (\ref{eq:training2}) is applicable for all trained networks, regardless the training algorithm used. Besides, with a bit modification, we can use (\ref{eq:training2}) to get the test set error of faulty RBF networks.
In (\ref{eq:training2}), the term $\frac{P_{\beta}}{N}\sum\limits_{i=1}^N y_i^2$ can be seen as a constant with respect to the weight vector $\ibw$. Hence the training objective function can be defined as
\beq \label{eq:training2_new}
\psi\left(\ibw\right) = \frac{1}{N} \left\| \iby - \bA \ibw \right\|_2^2+ \ibw^\mathrm{T}\bR\ibw,
\eeq
where $\bR = (P_{\beta}+\sigma_b^2)\mbox{diag}\left(\frac{1}{N}\bA^\mathrm{T} \bA\right)-\frac{P_{\beta}}{N}\bA^\mathrm{T} \bA$.

\subsection{ADMM}
The ADMM framework is an iterative approach for solving optimization problems by breaking them into smaller subproblems \cite{Boyd2011}. This algorithm can be used to solve problems in the standard form
\begin{subequations}
\beq
& & \min\limits_{\ibz,\iby}:~\psi\left(\ibz\right) + g\left(\iby\right) \\
& & s.t. ~~~~\bC \ibz + \bD \iby = \ibb .
\eeq
\label{eq:admm_obj}
\end{subequations}
with variables $\ibz\in \mathbb{R}^n$ and $\iby\in \mathbb{R}^m$, where $\ibb\in\mathbb{R}^p$, $\bC \in \mathbb{R}^{p\times n}$ and $\bD \in \mathbb{R}^{p\times m}$.
In the ADMM framework, first we construct an augmented Lagrangian function as
\beq
&L \left(\ibz, \iby, \bal \right) = & \psi\left(\ibz\right) + g\left(\iby\right) + \bal^\mathrm{T} \left( \bC \ibz + \bD \iby  - \ibb \right) \nonumber\\
& &+ \frac{\rho}{2} \left\|\bC \ibz + \bD \iby  - \ibb \right\|_2^2,
\label{eq:admm_augmented}
\eeq
where $\bal\in\mathbb{R}^p$ is the Lagrange multiplier vector, and $\rho>0$.
The algorithm consists of the iterations as:
\begin{subequations}
\label{eq:admm_scheme}
\beq
\ibz^{k+1} &=& \argmin\limits_{\iby} L(\ibz^{k}, \iby, \bal^k),\\
\iby^{k+1} &=& \argmin\limits_{\ibz} L(\ibz, \iby^{k+1}, \bal^k),\\
\bal^{k+1} &=& \bal^k + \rho \left( \bC \ibz^{k+1} + \bD \iby^{k+1} - \ibb \right).
\eeq
\end{subequations}
For updating the dual variable, a step size equal to the augmented Lagrangian parameter $\rho$ is used.
\begin{figure}[h]
\centering
\centerline{\includegraphics[height=2in]{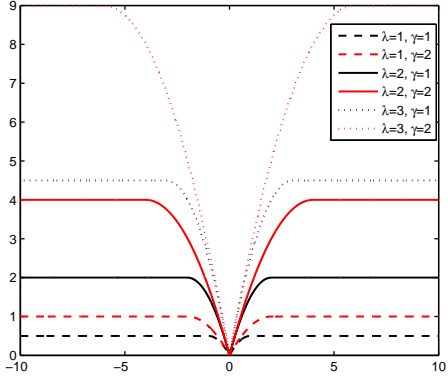}}
\caption{The shapes of MCP penalty function under different parameter settings.}
\label{fig:MCP}
\end{figure}

\section{Development of Proposed Algorithm} \label{section3}
In \eqref{eq:training2_new}, we use $M$ RBF centers. To limit the size of the RBF network and automatically select appropriate centers during training process, we further modify the  objective function and propose two novel algorithms based on the property of $l_0$-norm and ADMM framework.

\subsection{RBF center selection based on ADMM-MCP}
In the first method, we consider introducing an additional penalty term $\lambda\|\ibw\|_0$ into \eqref{eq:training2_new}, then we have
\beq\label{problem1_0}
\hat{Q}(\ibw,\blam)=\frac{1}{N}\|\iby-\bA\ibw\|_2^2+\ibw^\mathrm{T}\bR\ibw+\lambda\|\ibw\|_0,
\eeq
where $\|\ibw\|_0$ represents the number of non-zero entries in $\ibw$. Due to the nature of $l_0$-norm, the problem in \eqref{problem1_0} is NP hard \cite{donoho2006compressed}.
To handle this issue, the MCP function is introduced, which is a very attractive approximate function of $l_0$-norm \cite{zhang2010nearly, breheny2011coordinate}.
Thus, the function in \eqref{problem1_0} can be rewritten as
\beq\label{problem1}
Q(\ibw,\blam)=\frac{1}{N}\|\iby-\bA\ibw\|_2^2+\ibw^\mathrm{T}\bR\ibw+ P_{\lambda,\gamma}(\ibw),
\eeq
where $P_{\lambda,\gamma}(\ibw)=\sum_{i=1}^M P_{\lambda,\gamma}(w_i)$  $(\lambda>0, \gamma>0)$ denotes the MCP function,
\beq
P_{\lambda,\gamma}(w_i) = \left\{ \begin{array}{lcl}
\lambda |w_i|-\frac{w_i^2}{2\gamma}, & if\,\,\,\, |w_i| \leq \gamma\lambda, \\
\frac{1}{2}\gamma\lambda^2, & if\,\,\,\, |w_i| > \gamma\lambda.
\end{array}\right.
\label{MCP}
\eeq
and
\beq
\frac{\partial P_{\lambda,\gamma}(\ibw)}{\partial w_i} &=& \lambda sign(w_i)\left(1-\frac{|w_i|}{\lambda\gamma}\right)_+ \nonumber\\
&=&\left\{ \begin{array}{lcl}
\lambda sign(w_i)-\frac{w_i}{\gamma}, & if\,\,\,\, |w_i| \leq \gamma\lambda, \\
0, & if\,\,\,\, |w_i| > \gamma\lambda.
\end{array}\right.
\eeq
The shapes of MCP penalty function with different parameter settings are shown in Fig.~\ref{fig:MCP}. From this figure, we see that, with appropriate parameter setting, the shape of MCP is very similar with the $l_0$-norm term.

Then the ADMM framework is used to handle the problem in \eqref{problem1}. Following the steps of ADMM, we introduce a dummy variable $\ibu=[u_1,\dots,u_M]^\mathrm{T}$ and transform the unconstrained problem into the standard constrained form
\begin{subequations}\label{meth1_orig}
\beq
&\min\limits_{\ibw,\ibu} &\psi(\ibw)+ P_{\lambda,\gamma}(\ibu), \\
&s.t. &\ibu=\ibw,
\eeq
\end{subequations}
where $\psi(\ibw)$ is given by \eqref{eq:training2_new}. Then we construct its augmented Lagrangian as
\beq\label{eq:Lagr_meth1}
L(\ibw, \ibu, \bupsi)&=&\psi(\ibw)+ P_{\lambda,\gamma}(\ibu)+\bupsi^\mathrm{T}(\ibu-\ibw) \nonumber\\
& &+ \frac{\rho}{2} \left\|\ibw - \ibu\right\|_2^2,
\eeq
According to \eqref{eq:admm_scheme}, the ADMM iteration of $\ibu^{k+1}$ is
\beq\label{meth1_u}
\ibu^{k+1}&=&\argmin\limits_{\ibu} L(\ibw^k, \ibu, \bupsi^k), \nonumber \\
&=&\small\argmin\limits_{\ibu} P_{\lambda,\gamma}(\ibu)+{\bupsi^k}^\mathrm{T}(\ibu-\ibw^k) +\frac{\rho}{2} \left\|\ibw^k - \ibu\right\|_2^2 \nonumber\\
&=&\small\argmin\limits_{\ibu} P_{\lambda,\gamma}(\ibu) + \frac{\rho}{2} \left\|\ibw^k-\ibu-\frac{1}{\rho}\bupsi^k \right\|_2^2
\eeq
where $\ibu^{k+1}=[u_1^{k+1},\dots,u_M^{k+1}]^\mathrm{T}$. For any $u_i$ ($i\in[1,\dots,{M}]$), if $\rho>\frac{1}{\gamma}$ then
\beq \label{meth2_ui_mcp1}
u_i^{k+1}=\left\{ \begin{array}{lcl}
\displaystyle \frac{S\left( w_i^k - \upsilon_i^k/\rho,\frac{\lambda}{\rho} \right)}{1-1/ (\gamma\rho)}, & if |w_i^k - \upsilon_i^k/\rho| \leq \gamma\lambda, \\
\displaystyle w_i^k - \upsilon_i^k/\rho, & if |w_i^k - \upsilon_i^k/\rho| > \gamma\lambda,
\end{array}\right.
\eeq
where $S$ denotes the soft-threshold operator \cite{donoho1994ideal} given by
\beq \label{eq:softThreshold}
S(z,\eta)=sign(z)\max\{|z|-\eta,0\}, \nonumber
\eeq
if $\rho=\frac{1}{\gamma}$,
\beq \label{meth2_ui_mcp2}
u_i^{k+1}=\left\{ \begin{array}{lcl}
\displaystyle 0, & if |w_i^k - \upsilon_i^k/\rho| \leq \gamma\lambda, \\
\displaystyle w_i^k - \upsilon_i^k/\rho, & if |w_i^k - \upsilon_i^k/\rho| > \gamma\lambda,
\end{array}\right.
\eeq
if $\rho<\frac{1}{\gamma}$,
\beq \label{meth2_ui_mcp3}
u_i^{k+1}=\left\{ \begin{array}{lcl}
\displaystyle 0, & if |w_i^k - \upsilon_i^k/\rho| \leq \sqrt{\frac{\gamma}{\rho}}\lambda, \\
\displaystyle w_i^k - \upsilon_i^k/\rho, & if |w_i^k - \upsilon_i^k/\rho| > \sqrt{\frac{\gamma}{\rho}}\lambda.
\end{array}\right.
\eeq
All these three cases have a unified approximate solution \cite{zhang2010nearly,breheny2011coordinate}
\beq \label{meth1_ui}
u_i^{k+1}=\left\{ \begin{array}{lcl}
\displaystyle \frac{S\left( w_i^k - \upsilon_i^k/\rho,\lambda \right)}{1-1/\gamma}, & if\,\,\,\, |w_i^k - \upsilon_i^k/\rho| \leq \gamma\lambda, \\
\displaystyle w_i^k-\upsilon_i^k/\rho, & if\,\,\,\, |w_i^k - \upsilon_i^k/\rho| > \gamma\lambda.
\end{array}\right.
\eeq
It is worth noting that when $\gamma\to\infty$ the function in \eqref{meth1_ui} is similar with the soft-threshold, when $\gamma\to 1$ the function is close to the hard-threshold.

$\ibw^{k+1}$ is directly calculated by first-order optimization condition, the solution is given by
\beq\label{meth1_w}
\ibw^{k+1} &=& \argmin\limits_{\ibw} L(\ibw, \ibu^{k+1}, \bupsi^k) \nonumber \\
&=&\small\argmin\limits_{\ibw} \psi(\ibw)+{\bupsi^k}^\mathrm{T}(\ibu^{k+1}-\ibw) + \frac{\rho}{2}\left\|\ibw-\ibu^{k+1}\right\|_2^2 \nonumber \\
&=&\argmin\limits_{\ibw} \psi(\ibw)+ \frac{\rho}{2} \left\|\ibw-\ibu^{k+1}-\frac{1}{\rho}\bupsi^k \right\|_2^2 \nonumber \\
&=&\small\left[\frac{2}{N}\bA^\mathrm{T}\bA+2\bR+\rho\bI\right]^{-1} \left[\frac{2}{N}\bA^\mathrm{T}\iby+\rho\ibu^{k+1}+\bupsi^k\right].
\eeq

$\bupsi^{k+1}$ is updated as
\beq\label{meth1_upsi}
\bupsi^{k+1} &=& \bupsi^k + \rho \left(\ibu^{k+1}-\ibw^{k+1}\right).
\eeq

\subsection{RBF center selection based on ADMM-HT}
In the first method, two parameters $\lambda$ and $\gamma$ are used.
%The settings of $\lambda$ and $\gamma$ are difficult when the magnitude of training set error is inconstant.
If we need the number of centers exactly equal to a certain value, we have to regularize $\lambda$ to meet this requirement. It is very inconvenient.
Hence we propose the second method which can directly limit the maximum number of centers.
First, the problem in \eqref{eq:training2_new} is modified as a constrained form
\begin{subequations}\label{problem2_0}
\beq
&\argmin\limits_{\ibw} &
\frac{1}{N}\left\|\iby-\bA\ibw\right\|_2^2+ \ibw^\mathrm{T}\bR\ibw, \\
& s.t. & \|\ibw\|_0\leq K.
\eeq
\end{subequations}
With the constraint in (\ref{problem2_0}b), we can ensure that the number of RBF centers is equal to or smaller than $K$.
But this problem cannot be directly solved by the ADMM framework.
Since the constraint in (\ref{problem2_0}b) is undecomposable.
To solve this issue, we introduce an indicator function
\beq \label{indicator}
i_{\ibc(K)}(\ibw) = \left\{\begin{array}{ll}
0, &if\,\,\,\,\ibw\in \ibc(K), \\
+\infty, &otherwise, \\
\end{array}\right.
\eeq
where the set $\ibc(K)=\{\ibw:\|\ibw\|_0\leq K\}$ ($K\leq M$). After that, the problem in \eqref{problem2_0} can be rewritten as
\beq \label{problem2}
\argmin\limits_{\ibw} \frac{1}{N}\left\|\iby-\bA\ibw\right\|_2^2 + \ibw^\mathrm{T}\bR\ibw + i_{\ibc(K)}(\ibw).
\eeq
Then, we introduce a dummy variable $\ibu$ and rewrite problem in \eqref{problem2} as
\begin{subequations}\label{meth2_orig}
\beq
&\argmin\limits_{\ibw} &\psi(\ibw) + g(\ibu), \\
& s.t. &\ibw=\ibu,
\eeq
\end{subequations}
where $g(\ibu)$ denotes the indicator function $i_{\ibc(K)}(\ibu)$. Its augmented Lagrangian is
\beq\label{eq:Lagr_meth2}
L(\ibw, \ibu, \bupsi)&=&\psi(\ibw)+g(\ibu)+\bupsi^\mathrm{T}(\ibu-\ibw) \nonumber\\
& &+ \frac{\rho}{2} \left\|\ibw - \ibu\right\|_2^2,
\eeq
According to \eqref{eq:Lagr_meth2} and \eqref{eq:admm_scheme}, we have:
\beq\label{meth2_u}
\ibu^{k+1} &=& \argmin\limits_{\ibu} L(\ibw^k, \ibu, \bupsi^k), \nonumber \\
&=& \argmin\limits_{\ibu} g(\ibu) + \frac{\rho}{2} \left\|\ibw^k-\ibu-\frac{1}{\rho}\bupsi^k \right\|_2^2,\nonumber \\
&=& HT\left(\ibw^k - \bupsi^k/\rho\right),
\eeq
where $HT(\ibz)$ sets all elements of $\ibz$ equal to 0 except the $K$ entries with the largest magnitudes. Obviously, the HT operation can restrict $\ibu$ into the feasible region $\ibc(K)$. It is worth noting that the second method just replaces the MCP function in the Lagrangian \eqref{eq:Lagr_meth1} by the indicator function $i_{\ibc(K)}(\ibu)$. It does not influence the update of $\ibw^{k+1}$ and $\bupsi^{k+1}$. Hence in the second method we still have:
\beq\label{meth2_w}
\small\ibw^{k+1}=\left[\frac{2}{N}\bA^\mathrm{T}\bA+2\bR+\rho\bI\right]^{-1} \left[\frac{2}{N}\bA^\mathrm{T}\iby+\rho\ibu^{k+1}+\bupsi^k\right].
\eeq
and
\beq\label{meth2_upsi}
\bupsi^{k+1} &=& \bupsi^k + \rho \left(\ibu^{k+1}-\ibw^{k+1}\right).
\eeq
\section{Analysis of Global Convergence} \label{section4}
In this section, we discuss the convergence of the proposed methods.
We cannot directly follow the general convergence proof for nonconvex ADMM given by \cite{wang2015global}.
Because some of the assumptions given in \cite{wang2015global} are not satisfied. For instance, $i_{\ibc(K)}(\ibu)$ is not a restricted prox-regular function. But the global convergence of our proposed two methods still can be proved. We first give a sketch of the proof in Theorem~\ref{the:CS_Convergence}, and the details are discussed latter.

\begin{theorem} \label{the:CS_Convergence}
{\it
If the proposed methods satisfy the following four conditions:

\noindent \textbf{C1} (Sufficient decrease condition)
For each $k$, $\exists \tau_1>0$ let
\beq
\small L(\ibw^{k+1}, \ibu^{k+1}, \bupsi^{k+1})-L(\ibw^{k}, \ibu^{k}, \bupsi^{k}) \leq -\tau_1\|\ibw^{k+1}-\ibw^{k}\|_2^2.
\eeq

\noindent \textbf{C2} (Boundness condition)
The sequences $\{\ibw^{k}, \ibu^{k}, \bupsi^{k}\}$ generated by the proposed two methods are bounded and their Lagrangian functions are lower bounded.

\noindent \textbf{C3} (Subgradient bound condition) For each $k\in \mathbb{N}$, there exists a
$\ibd^{k+1}\in \partial L(\ibw^{k+1},\ibu^{k+1},\bupsi^{k+1})$, and a $\tau_2>0$ such that
\beq
\|\ibd^{k+1}\|_2^2\leq \tau_2\|\ibw^{k+1}-\ibw^{k}\|_2^2.
\eeq

\noindent \textbf{C4} (Continuity condition) If $\{\ibw^{*}, \ibu^{*}, \bupsi^{*}\}$ is the limit point of a sub-sequence $\{\ibw^{k_s}, \ibu^{k_s}, \bupsi^{k_s}\}$ $( s\in \mathbb{N})$, then $L(\ibw^{*},\ibu^{*},\bupsi^{*})=\lim_{s \to \infty} L(\ibw^{k_s},\ibu^{k_s},\bupsi^{k_s})$.

\noindent Based on these conditions, we can further deduce that the sequences $\{\ibw^{k}, \ibu^{k}, \bupsi^{k}\}$ have at least one limit point $\{\ibw^{*}, \ibu^{*}, \bupsi^{*}\}$ and any limit point $\{\ibw^{*}, \ibu^{*}, \bupsi^{*}\}$ is a stationary point. Moreover, if their Lagrangian functions $L(\ibw,\ibu,\bupsi)$ are Kurdyka-{\L}ojasiewicz (K{\L}) function, then the sequences $\{\ibw^{k}, \ibu^{k}, \bupsi^{k}\}$ will globally converge to a unique limit point $\{\ibw^{*}, \ibu^{*}, \bupsi^{*}\}$.}
\end{theorem}

\textbf{Proof:} The theorem is similar as Proposition 2 in \cite{wang2015global} and Theorem 2.9 in \cite{attouch2013convergence}. The proof of it is also standard. So we omit it here. The details can be found in the proof of Proposition 2 in \cite{wang2015global}. $\blacksquare$

For the proof of convergence, the key step here is to prove that the above mentioned four conditions are satisfied. Hence, we have the following four propositions.

\begin{proposition} \label{prop:C1}
{\it
If $\rho$ is greater than a certain value, the proposed two methods satisfy the sufficient decrease condition in \textbf{C1}.}
\end{proposition}

\textbf{Proof:} In the following proof, we use the second method as an example. For the first method, the proof is same except replacing the function $g(\ibu)$ by $P_{\lambda,\gamma}(\ibu)$.

For the second method, the Lagrangian function in \eqref{eq:Lagr_meth2} can be rewritten as
\beq\label{eq:Lagr_meth2_1}
L(\ibw, \ibu, \bupsi)&=&\psi(\ibw) + \frac{\rho}{2} \left\|\ibw-\ibu-\frac{1}{\rho}\bupsi \right\|_2^2 \nonumber \\ &&+g(\ibu) -\frac{1}{2\rho}\|\bupsi\|_2^2.
\eeq
The second-order derivative of $\bpsi(\ibw)$ is
\beq \label{2nd_grad}
\frac{\partial^2\bpsi(\ibw)}{\partial \ibw^2}&=&\frac{2}{N}\bA^\mathrm{T}\bA+2\bR \\
&=&\frac{2}{N}[(1-P_{\beta})\bA^\mathrm{T}\bA +(P_{\beta}+\sigma_b^2)diag(\bA^\mathrm{T}\bA)] \nonumber.
\eeq
Obviously, it is positive definite. Hence $\bpsi(\ibw)$ is strongly convex.
We can further deduce that the Lagrangian in \eqref{eq:Lagr_meth2_1} is also strongly convex with respect to $\ibw$. Hence, based on the definition of strongly convex function, we have
\beq \label{eq:1st}
L(\ibw^{k+1}, \ibu^{k+1}, \bupsi^{k})-L(\ibw^{k}, \ibu^{k+1}, \bupsi^{k}) \nonumber \\ \leq-\frac{a}{2}\|\ibw^{k+1}-\ibw^{k}\|_2^2,
\eeq
where $a>0$.

From \eqref{meth2_w}, we have
\beq
\nabla\psi(\ibw^{k+1})-\bupsi^{k} +\rho(\ibw^{k+1}-\ibu^{k+1})=0.
\eeq
Combining it with \eqref{meth2_upsi}, we can deduce that $\nabla\psi(\ibw^{k+1})=\bupsi^{k+1}$ and $\ibu^{k+1}-\ibw^{k+1} = 1/\rho\left(\bupsi^{k+1}-\bupsi^k\right)$. Thus
\beq \label{eq:2nd}
&&L(\ibw^{k+1}, \ibu^{k+1}, \bupsi^{k+1})-L(\ibw^{k+1}, \ibu^{k+1}, \bupsi^{k}) \nonumber \\
&=& \left(\bupsi^{k+1}-\bupsi^{k}\right)^\mathrm{T}\left(\ibu^{k+1}-\ibw^{k+1}\right) \nonumber \\ &=&\frac{1}{\rho}\|\bupsi^{k+1}-\bupsi^{k}\|_2^2
=\frac{1}{\rho}\|\nabla\psi(\ibw^{k+1})-\nabla\psi(\ibw^{k})\|_2^2 \nonumber \\
&\leq & \frac{l_{\psi}^2}{\rho}\|\ibw^{k+1}-\ibw^{k}\|_2^2,
\eeq
where $l_{\psi}$ is a Lipschitz constant of function $\psi(\ibw)$. The last inequality in \eqref{eq:2nd} is because that $\psi(\ibw)$ has Lipschitz continuous gradient.

From \eqref{meth2_u}, we have
\beq \label{eq:3rd}
L(\ibw^{k}, \ibu^{k+1}, \bupsi^{k})-L(\ibw^{k}, \ibu^{k}, \bupsi^{k})\leq 0
\eeq
Combining \eqref{eq:1st}, \eqref{eq:2nd} and \eqref{eq:3rd}, we have
\beq \label{eq:suff_desc}
&&L(\ibw^{k+1}, \ibu^{k+1}, \bupsi^{k+1})-L(\ibw^{k}, \ibu^{k}, \bupsi^{k}) \nonumber \\
&=&L(\ibw^{k+1}, \ibu^{k+1}, \bupsi^{k+1})-L(\ibw^{k+1}, \ibu^{k+1}, \bupsi^{k}) \nonumber \\
&&+L(\ibw^{k+1}, \ibu^{k+1}, \bupsi^{k})-L(\ibw^{k}, \ibu^{k+1}, \bupsi^{k}) \nonumber \\
&&+L(\ibw^{k}, \ibu^{k+1}, \bupsi^{k})-L(\ibw^{k}, \ibu^{k}, \bupsi^{k}) \nonumber \\
&\leq& \left(\frac{l_{\psi}^2}{\rho}-\frac{a}{2}\right)\|\ibw^{k+1}-\ibw^{k}\|_2^2.
\eeq
If $\rho>2l_{\psi}^2/a$, we can ensure that $l_{\psi}^2/\rho-a/2<0$. Hence the $\tau_1=a/2-l_{\psi}^2/\rho$ in \textbf{C1}. $\blacksquare$

\begin{proposition} \label{prop:C2}
{\it
If $\rho\geq l_{\psi}$ , $L(\ibw^{k}, \ibu^{k}, \bupsi^{k})$ is lower bounded, and the sequences $\{\ibw^{k}, \ibu^{k}, \bupsi^{k}\}$ generated by the proposed methods are bounded.}
\end{proposition}

\textbf{Proof:} We still use the second method as an example. The proof of the first method is also similar.
Firstly, we prove the $L(\ibw^{k}, \ibu^{k}, \bupsi^{k})$ is lower bounded for all $k$.
\beq\label{eq:Lagr_meth2_2}
L(\ibw^k, \ibu^k, \bupsi^k)=\psi(\ibw^k)+g(\ibu^k)+{\bupsi^k}^\mathrm{T}(\ibu^k-\ibw^k) \nonumber\\
+ \frac{\rho}{2} \left\|\ibw^k - \ibu^k\right\|_2^2, \nonumber\\
=\psi(\ibw^k)+g(\ibu^k)+{\nabla\psi(\ibw^{k})}^\mathrm{T}(\ibu^k-\ibw^k) \nonumber\\
+ \frac{\rho}{2} \left\|\ibw^k - \ibu^k\right\|_2^2, \nonumber\\
\geq \psi(\ibu^k)+\left(\frac{\rho}{2}-\frac{l_{\psi}}{2}\right)\|\ibu^k-\ibw^k\|_2^2 +g(\ibu^k),
\eeq
where the inequality in \eqref{eq:Lagr_meth2_2} is due to Lemma 3.1 in \cite{attouch2013convergence} and the Lipschitz continue gradient of $\psi(\ibw)$. According to Lemma 3.1 in \cite{attouch2013convergence}, we can deduce that
\beq
\psi(\ibw^k)+{\nabla\psi(\ibw^{k})}^\mathrm{T}(\ibu^k-\ibw^k) \geq \psi(\ibu^k) -\frac{l_{\psi}}{2}\|\ibu^k-\ibw^k\|_2^2, \nonumber
\eeq
thus we have the inequality in \eqref{eq:Lagr_meth2_2}.

Obviously, if $\rho\geq l_{\psi}$, then $\psi(\ibu^k)+\left(\frac{\rho}{2}-\frac{l_{\psi}}{2}\right)\|\ibu^k-\ibw^k\|_2^2 +g(\ibu^k)>-\infty$ and $L(\ibw^k, \ibu^k, \bupsi^k)$ is lower bounded.
According to the proof of Property 1, we know $L(\ibw^k, \ibu^k, \bupsi^k)$ is sufficient descent.
Hence $L(\ibw^k, \ibu^k, \bupsi^k)$ is upper bounded by $L(\ibw^0, \ibu^0, \bupsi^0)$.
%Thus, we can say that $L(\ibw^k, \ibu^k, \bupsi^k)$ is bounded.
%Due to the sufficient descent property, $L(\ibw^{k}, \ibu^{k}, \bupsi^{k})$ must converge when $k\to\infty$ and as long as $\ibu^0\in\ibc(K)$, for $\forall k$, we have $\ibu^k\in\ibc(K)$.

Next, we prove the sequence $\{\ibw^{k}, \ibu^{k}, \bupsi^{k}\}$ is bounded. From inequation \eqref{eq:suff_desc}, we have
\beq \label{eq:wk1_wk}
&&\|\ibw^{k+1}-\ibw^k\|_2^2\nonumber \\ &&\leq\frac{1}{\tau_1}\left(L(\ibw^k, \ibu^k, \bupsi^k)- L(\ibw^{k+1}, \ibu^{k+1}, \bupsi^{k+1})\right). \nonumber
\eeq
Then, we can deduce that
\beq \label{eq:wk1_wk}
&& \sum_{k=1}^{l} \|\ibw^{k+1}-\ibw^k\|_2^2\nonumber \\ &&\leq\frac{1}{\tau_1}\left(L(\ibw^0, \ibu^0, \bupsi^0)- L(\ibw^{l+1}, \ibu^{l+1}, \bupsi^{l+1})\right) \nonumber \\
&&< \infty.
\eeq
If $l\to\infty$, we still have $\sum_{k=1}^{\infty}\|\ibw^{k+1}-\ibw^k\|_2^2 < \infty$. Thus $\ibw^{k}$ is bounded.

From \eqref{eq:2nd}, we know
\beq
\|\bupsi^{k+1}-\bupsi^{k}\|_2^2 \leq l_{\psi}^2 \|\ibw^{k+1}-\ibw^{k}\|_2^2. \nonumber
\eeq
Therefore we can also deduce that
\beq
\sum_{i=1}^{\infty} \|\bupsi^{k+1}-\bupsi^k\|_2^2<\infty.
\eeq

In addation, according to \eqref{meth2_upsi}, we have
\beq
&&\|\ibu^{k+1}-\ibu^k\|_2^2\nonumber \\
&=&\|\ibw^{k+1}-\ibw^k+\frac{1}{\rho} \left(\bupsi^{k+1}-\bupsi^k\right)+\frac{1}{\rho} \left(\bupsi^{k-1}-\bupsi^k\right)\|_2^2 \nonumber \\
&\leq& 2\|\ibw^{k+1}-\ibw^k\|_2^2+\frac{2}{\rho^2}\|\bupsi^{k+1}-\bupsi^k\|_2^2\nonumber\\ &&+\frac{2}{\rho^2}\|\bupsi^{k-1}-\bupsi^k\|_2^2.
\eeq
Thus
\beq
\sum_{i=1}^{\infty} \|\ibu^{k+1}-\ibu^k\|_2^2 < \infty.
\eeq
So the sequence $\{\ibw^{k}, \ibu^{k}, \bupsi^{k}\}$ is bounded. $\blacksquare$

\begin{proposition} \label{prop:C3}
{\it
The proposed two methods satisfy the subgradient bound condition given by \textbf{C3}.}
\end{proposition}

\textbf{Proof:} For the second method,
\beq
&&\left. \frac{\partial L}{\partial \ibw} \right|_{(\ibw^{k+1}, \ibu^{k+1}, \bupsi^{k+1})} \nonumber \\
&=&\nabla \psi(\ibw^{k+1})+\rho\left(\ibw^{k+1}-\ibu^{k+1}\right)-\bupsi^{k+1} \nonumber \\
&=&\bupsi^{k}-\bupsi^{k+1}, \label{eq:dldw}\\
&&\left. \frac{\partial L}{\partial \ibu} \right|_{(\ibw^{k+1}, \ibu^{k+1}, \bupsi^{k+1})} \nonumber \\
&=&\partial g(\ibu^{k+1})-\rho\left(\ibw^{k+1}-\ibu^{k+1}\right)+\bupsi^{k+1} \nonumber \\
&\owns&\rho\left(\ibw^{k}-\ibw^{k+1}\right)+\bupsi^{k+1}-\bupsi^{k}, \label{eq:dldu}\\
&&\left. \frac{\partial L}{\partial \bupsi} \right|_{(\ibw^{k+1}, \ibu^{k+1}, \bupsi^{k+1})} \nonumber \\
&=&\ibu^{k+1}-\ibw^{k+1}=\frac{1}{\rho}\left(\bupsi^{k+1}-\bupsi^{k}\right), \label{eq:dldupsi}
\eeq
where the second equality in \eqref{eq:dldu} is according to \eqref{meth2_u},
\beq
0\in\partial g(\ibu^{k+1})+\bupsi^{k}-\rho\left(\ibw^k-\ibu^{k+1}\right),
\eeq
where $\partial$ is a generalized notion called limiting-subdifferential \cite{attouch2013convergence}. Being given $\ibu^{k+1}\in\ibc(K)$, the limiting-subdifferential of $g(\ibu)$ at $\ibu^{k+1}$ is called the normal cone to $\ibc(K)$ at $\ibu^{k+1}$, which is denoted by $N_\ibc(\ibu^{k+1})$ (for $\ibu^{k+1} \notin \ibc(K)$ we set $N_\ibc(\ibu^{k+1})= \emptyset$ ) \cite{attouch2013convergence}.
Thus
\beq
\ibd^{k+1}&:=&
\left[\begin{array}{c}
\bupsi^{k}-\bupsi^{k+1} \\
\rho\left(\ibw^{k}-\ibw^{k+1}\right)+\bupsi^{k+1}-\bupsi^{k}\\
\frac{1}{\rho}\left(\bupsi^{k+1}-\bupsi^{k}\right)
\end{array}\right] \nonumber\\
&\in&\partial L\left(\ibw^{k+1}, \ibu^{k+1}, \bupsi^{k+1}\right)
\eeq
Combining it with the inequality in \eqref{eq:2nd}, we can deduce that
\beq
\|\ibd^{k+1}\|_2^2 \leq \tau_2\|\ibw^{k+1}-\ibw^k\|_2^2.
\eeq
Apparently, for the first method, we just need to replace the function $g(\ibu)$ by $P_{\lambda,\gamma}(\ibu)$, then all other proof is same with the above mentioned process. $\blacksquare$

Then, for proving the condition in \textbf{C4}, the following lemma is introduced.

\begin{lemma} \label{lemma1}
{\it
The indicator function in \eqref{indicator} is low semi-continuous.
% if the set $\ibc(K)=\{\ibu:\|\ibu\|_0\leq K\}$ is closed for $\forall K\leq {M}$.
}
\end{lemma}

\textbf{Proof:}
From \cite{baire1905leccons}, we know that suppose $\bX$ is a topological space, a function $f:\,\,\bX\to \mathbb {R} \cup \{-\infty ,\infty \}$ is lower semi-continuous if and only if all of its lower levelsets $\{\ibx\in \bX:\,\,f(\ibx)\leq \alpha \}$  are closed for every $\alpha$. %\cite{baire1905leccons}.

According to this property, to prove Lemma~\ref{lemma1}, we just need to prove that for every $\alpha$, the set $g(\ibu)\leq \alpha$ is closed.
Obviously, the set is $\emptyset$ or $\ibc(K)$ for $\forall \alpha$.
Hence if $\ibc(K)=\{\ibu:\|\ibu\|_0\leq K\}$ is closed, then $g(\ibu)$ is a lower semi-continuous function.

Then, we prove that $\ibc(K)=\{\ibu:\|\ibu\|_0\leq K\}$ $(K\leq {M})$ is a closed set.
In other words, we need to prove that its complementary set $\ibc^c(K)=\{\ibu:\|\ibu\|_0>K\}$ $(K\leq {M})$ is an open set.
According to the definition of open set, for $\forall \ibu\in \ibc^c(K)$, $\ibx$ is an arbitrary variable in the neighbourhood of $\ibu$ where $\|\ibx-\ibu\|<\epsilon$ ($\epsilon>0$).
If $\epsilon$ is small enough then $\ibx\in \ibc^c(K)$. Hence the set $\ibc^c(K)=\{\ibu:\|\ibu\|_0> K\}$ is open, and the set $\ibc(K)=\{\ibu:\|\ibu\|_0\leq K\}$ $(K\leq {M})$ is closed.
Further more, $g(\ibu)$ is lower semi-continuous. $\blacksquare$

\begin{proposition} \label{prop:C4}
{\it
The proposed two methods satisfy the continuity condition in \textbf{C4}.}
\end{proposition}

\textbf{Proof:}
If the Lagrangian functions of our proposed methods are lower semi-continuous, the continuity condition in \textbf{C4} is satisfied \cite{wang2015global,attouch2013convergence}.
For the first method, the Lagrangian function is continuous. Thus, it must be a lower semi-continuous function.
For the second method, based on Lemma~\ref{lemma1}, we know that the indicator function $g(\ibu)$ is lower semi-continuous.
Besides, the terms $\psi(\ibw)$, $\bupsi^\mathrm{T}(\ibu-\ibw)$ and $\tilde{\rho}/2\|\ibw-\ibu\|_2^2$ are all continuous function. Then we can deduce that its Lagrangian function is lower semi-continuous. The condition in \textbf{C4} is satisfied.
$\blacksquare$

Thus, based on \textbf{C1}-\textbf{C4} in Theorem~\ref{the:CS_Convergence}, we know that the sequences $\{\ibw^{k}, \ibu^{k}, \bupsi^{k}\}$ generated by the first method and second method both have at least one limit point $\{\ibw^{*}, \ibu^{*}, \bupsi^{*}\}$ and any limit point $\{\ibw^{*}, \ibu^{*}, \bupsi^{*}\}$ is a stationary point.
%In other words, at least, the proposed two methods have local convergence.
Finally, to ensure that both of them have global convergence to a unique limit point, we need to prove that their Lagrangian functions are K{\L} function.
Before that, in order to facilitate the following explanation, we introduce several fundamental definitions.

For a function $f:\,\,\mathbb{R}^n\to\mathbb{R}$, $dom\,\,f$ denotes the domain of $f$. A function $f$ is proper means that $dom\,\,f\neq\emptyset$ and it can never attain $-\infty$.

%It is lower semi-continuous at a point $\ibx_0$ if
%\beq
%\liminf \limits_{\ibx\to\ibx_0}\,f(\ibx)\geq f(\ibx_0). \nonumber
%\eeq
%And if the function $f$ is lower semi-continuous at every point in $dom\,\,f$, then it is a lower semi-continuous function.

A subset $S$ of $\mathbb{R}^d$ is a real semi-algebraic set if there exists a finite number of real polynomial functions $l_{ij},\,\,h_{ij}: \mathbb{R}^d\to\mathbb{R}$ such that
\beq
S=\bigcup_{j=1}^{q_1}\bigcap_{i=1}^{q_2}\{\bz\in\mathbb{R}^d:\, l_{ij}(\bz)=0, \,h_{ij}(\bz)<0\}. \nonumber
\eeq
A function $h: \mathbb{R}^d\to(-\infty,\infty]$ is semi-algebraic if its graph
\beq
\{(\bz,t)\in\mathbb{R}^{d+1}:h(\bz)=t\}
\eeq
is a semi-algebraic set in $\mathbb{R}^{d+1}$.

The definition of K{\L} function is given by \cite{attouch2013convergence}. Here we use Lemma~\ref{lemma2} to prove that a function is K{\L} function.

\begin{lemma} \label{lemma2}
{\it
Let $f: \mathbb{R}^n\to(-\infty,\infty]$ be a proper and lower semi-continuous function. If f is semi-algebraic, then it satisfies the K{\L} property at any point of $dom\,\,f$. In other words, $f$ is a K{\L} function.}
\end{lemma}

\textbf{Proof:} The proof of Lemma~\ref{lemma2} is given by \cite{bolte2006nonsmooth}.
%bolte2007lojasiewicz
$\blacksquare$

Then, based on Lemma~\ref{lemma2}, we prove $L(\ibw^{k}, \ibu^{k}, \bupsi^{k})$ is a K{\L} function. Obviously, for each proposed method, the $L(\ibw^{k}, \ibu^{k}, \bupsi^{k})$ is a proper and lower semi-continuous function.
The key is to prove that $L(\ibw^{k}, \ibu^{k}, \bupsi^{k})$ is semi-algebraic.
All other terms are obvious, except $P_{\lambda,\gamma}(\ibu)$ in the first method and $g(\ibu)$ in the second method.
From \cite{zhang2016alternating}, we know that the MCP function $P_{\lambda,\gamma}(\ibu)$ is a semi-algebraic function.

For $g(\ibu)$, according to the definition of semi-algebraic function, we need to prove that its graph $\{(\ibu,t)\in\mathbb{R}^{{M}+1}:g(\ibu)=t\}$ is a semi-algebraic subset of $\mathbb{R}^{{M}+1}$.
If $t\neq 0$, the set $\{g(\ibu)=t\}$ is empty.
Hence, the graph $\{(\ibu,t)\in\mathbb{R}^{{M}+1}:g(\ibu)=t\}$ is equal to $\{\ibu\in\mathbb{R}^{{M}}:\,\,g(\ibu)=0\}$ which can be rewritten as $\ibc(K)=\{\ibu:\,\,\|\ibu\|_0\leq K\}=\bigcup_{k=1}^K\{\ibu:\,\,\|\ibu\|_0=k\}$.
We know that a semi-algebraic set is stable under finite union operation.
%Hence we need to prove that for $\forall k\leq K$, $\{\ibu:\|\ibu\|_0=k\}$ is semi-algebraic.
For $\forall k\leq K$, $\{\ibu:\|\ibu\|_0=k\}=\{\ibu=[u_1,\dots,u_{M}]^\mathrm{T}: \sum_{i=1}^{M} u_i^0=k\}$ is semi-algebraic. Therefore $\ibc(K)$ is a semi-algebraic set, and $g(\ibu)$ and $L(\ibw^{k}, \ibu^{k}, \bupsi^{k})$ are semi-algebraic functions.
Further, $L(\ibw^{k}, \ibu^{k}, \bupsi^{k})$ is a K{\L} function for the proposed two methods.

In summary, as long as $\tilde{\rho}>\max\{2\hat{l}_{\psi}^2/\tilde{a},\hat{l}_{\psi}\}$, both proposed methods have global convergence to a unique limit point.

\begin{table}[tbp]
\newcommand{\tabincell}[2]{\begin{tabular}{@{}#1@{}}#2\end{tabular}}
\centering
\begin{tabular}{|c|c|c|c|c|}
\hline
Dataset &\tabincell{c}{Number of \\features} &\tabincell{c}{Size of \\training set} &\tabincell{c}{Size of \\test set} &\tabincell{c}{RBF \\width}\\ \hline
Abalon &7 &2000 &2177 &0.1\\
Airfoil Self-Noise &5 &751 &752 &0.5\\
Boston Housing &13 &400 &106 &2\\
Concrete &9 &500 &530 &0.5\\
Energy Efficiency &7 &600 &168 &0.5\\
Wine Quality (white) &12 &2000 &2898 &1\\ \hline
\end{tabular}
\caption{Properties of six data sets.}
\label{table1}
\end{table}

\begin{figure*}[th]
\begin{tabular}{c@{\extracolsep{10mm}}c@{\extracolsep{10mm}}c}
\mbox{\epsfig{figure=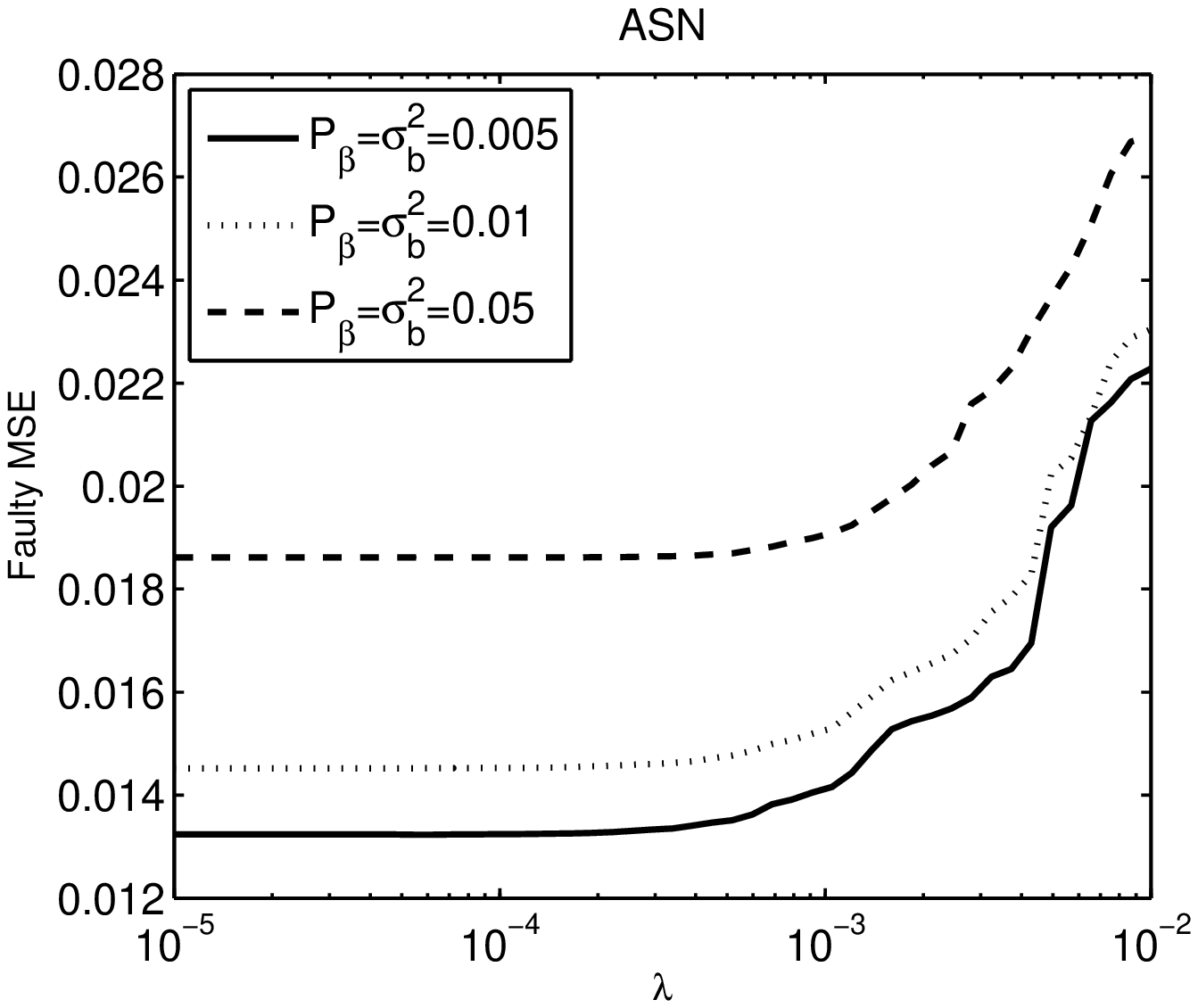,width=2in}}
&\mbox{\epsfig{figure=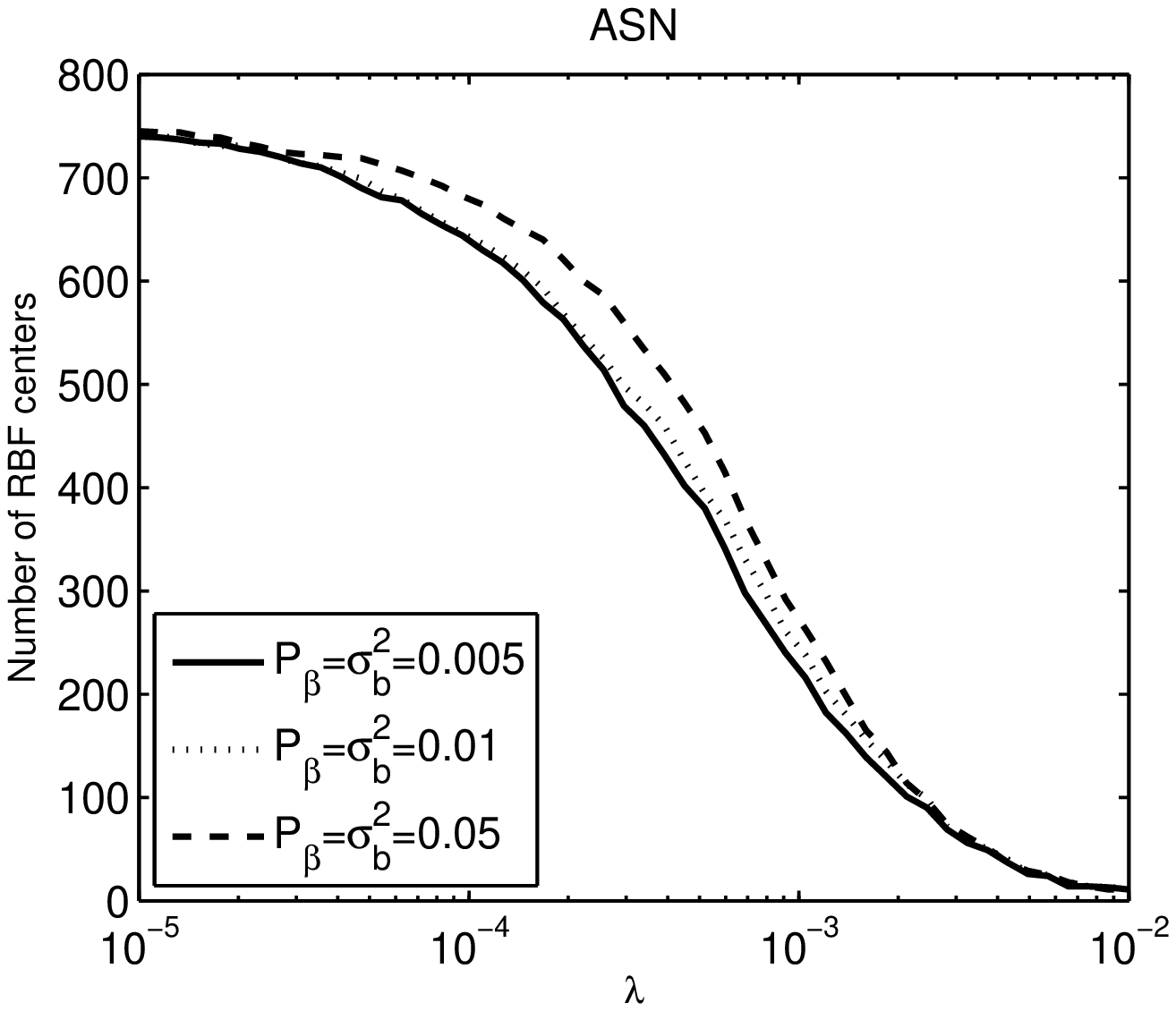,width=2in}}
&\mbox{\epsfig{figure=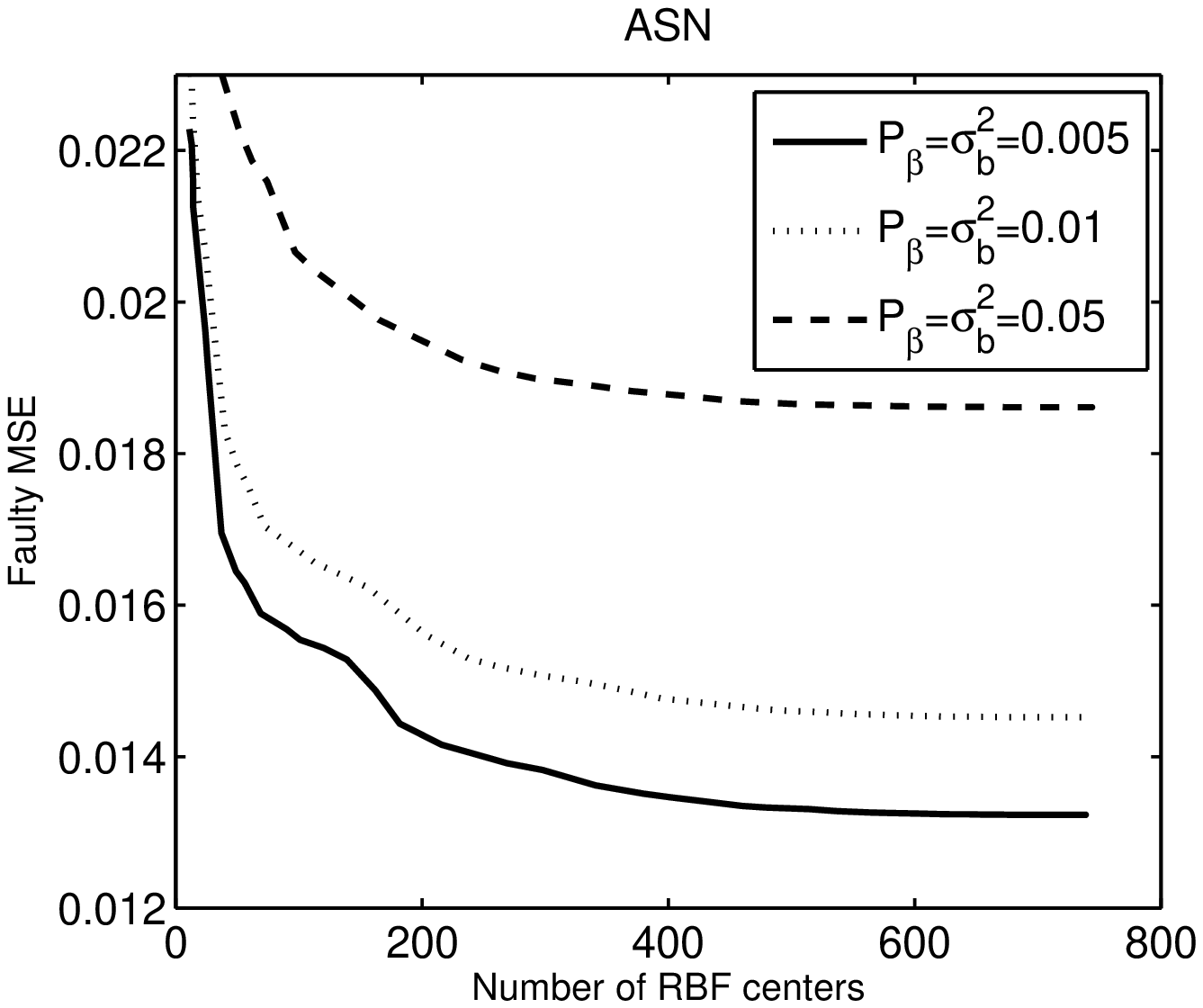,width=2in}}
\end{tabular}
\caption{Properties of ADMM-MCP.}
\label{fig:firstMethod}
\end{figure*}

%\begin{figure}[h]
%\centering
%%\begin{tabular}{c}
%%\mbox{\epsfig{figure=ASN_seed_1_mse_MCP_convex.eps,width=1.6in}} &
%\mbox{\epsfig{figure=ASN_seed_1_mse_MCP.eps,width=2in}}  \\
%%\mbox{\epsfig{figure=seed_1_num_lam_MCP_convex.eps,width=1.6in}} &
%\mbox{\epsfig{figure=seed_1_num_lam_MCP.eps,width=2in}} \\
%%\mbox{\epsfig{figure=ASN_seed_1_mse_num_MCP_convex.eps,width=1.6in}} &
%\mbox{\epsfig{figure=ASN_seed_1_mse_num_MCP.eps,width=2in}}
%%\end{tabular}
%\caption{Properties of ADMM-MCP.}
%\label{fig:firstMethod}
%\end{figure}

\begin{figure}[h]
\centering
\centerline{\includegraphics[width=2in]{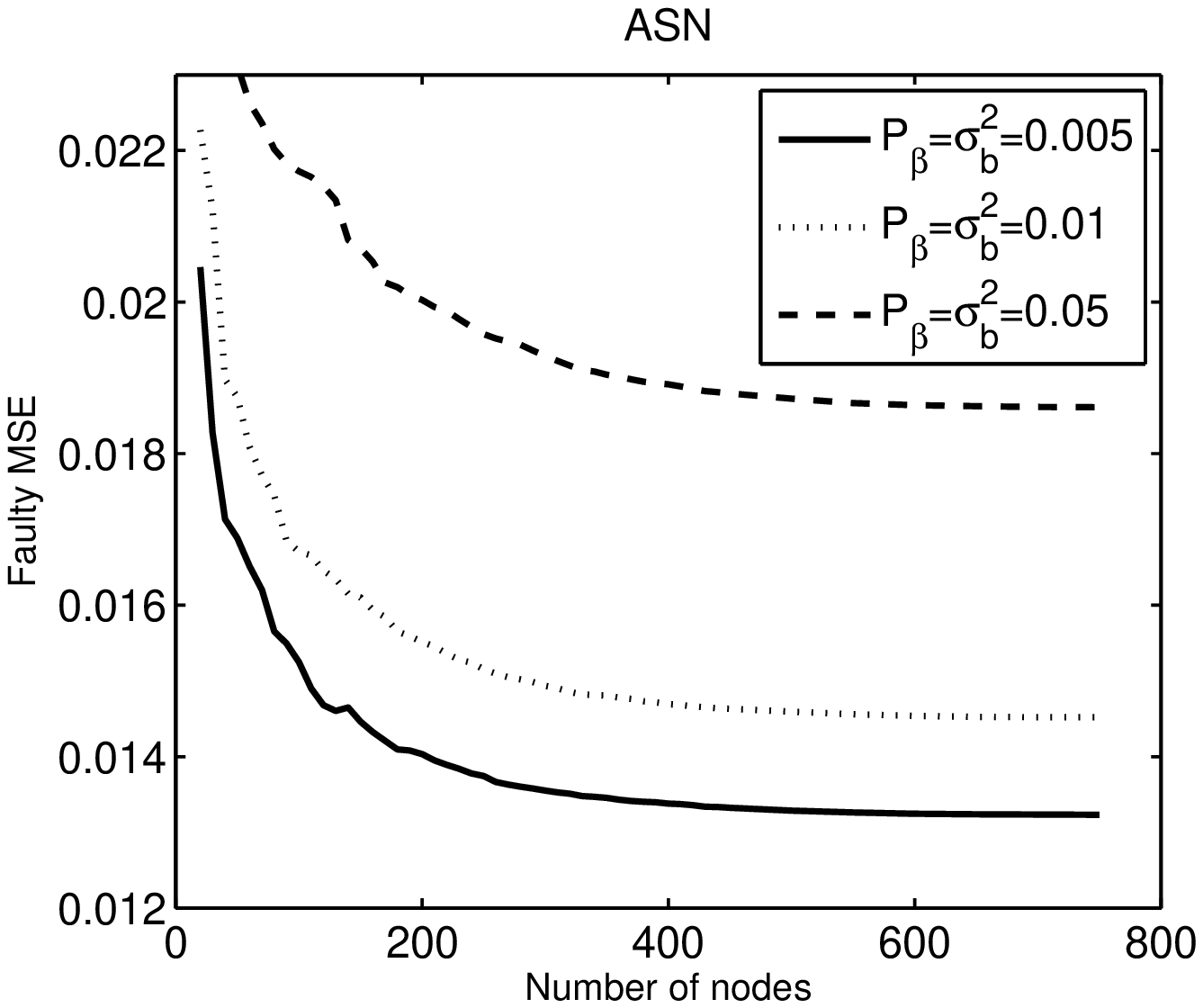}}
\caption{Properties of ADMM-HT.}
\label{fig:secondMethod}
\end{figure}

\section{Simulation Result} \label{section5}
\subsection{Settings}
In the following experiments, six University of California Irvine (UCI) regression datasets are used \cite{Lichman2013}. They are respectively Abalone (ABA)~\cite{zhang2015comparison,paper:sugiyama2002}, Airfoil Self-Noise (ASN)~\cite{li2016inverse}, Boston Housing (HOUSING)~\cite{zhang2015comparison,li2016inverse}, Concrete (CON)~\cite{le2014optimal}, Energy Efficiency (ENERGY)~\cite{li2016inverse}, and Wine Quality (white) (WQW)~\cite{ekambaram2016active,cortez2009modeling}. For each dataset, its RBF width is selected between 0.1 to 10. The basic setting of each dataset is given by TABLE~\ref{table1}.

In the following experiment, the performance of all algorithms is evaluated by the average test set error
\beq \label{eq:testerror}
\overline{\cal E}_{test} = \frac{P_{\beta}}{N'}\sum\limits_{i'=1}^{N'} {y'}_{i'}^2 +
\frac{1-P_{\beta}}{N'} \left\| \iby' - \bA' \ibw \right\|_2^2 \nonumber\\
+ \frac{1-P_{\beta}}{N'}\ibw^\mathrm{T}\left[(P_{\beta}+\sigma_b^2)\mbox{diag} \left({\bA'}^\mathrm{T} \bA'\right)-P_{\beta}{\bA'}^\mathrm{T} \bA'\right]\ibw,
\eeq
where $N'$ denotes the size of test set, $\iby'=[{y'}_{1},\dots,{y'}_{N'}]$, $\bA'$ is a $N'\times M$ matrix, and its element in $i’$th row and $j$th column is
\beq
[\bA']_{i',j}=\exp\left(-\frac{\left\|{\ibx'}_{i'}-\ibc_j\right\|_2^2}{s}\right). \nonumber
\eeq
%Besides, $P_{\beta}$ and $\sigma_b^2$ are two parameters which can be used to control the level of open weight fault and multiplicative weight fault respectively.
%In our experiments, we consider three fault scenarios: $\{P_\beta=\sigma_b^2=0.005\}$, $\{P_\beta=\sigma_b^2=0.01\}$ and $\{P_\beta=\sigma_b^2=0.05\}$.

To better approximate the $l_0$-norm term, the first method uses the MCP function with $\gamma=1.001$.
%And it can further separated into two cases based on whether the parameter $\gamma$ satisfies the condition in Theorem~\ref{the:CS_Convexity}.
%If it is satisfied ($\gamma=1/c^*+0.001$ in our experiments), we call it convex ADMM-MCP method.
%Otherwise, it is a general ADMM-MCP method.
%For the second case, we let $\gamma=1.001$.
The parameter $\lambda$ is used to control the number of nodes.
The corresponding experimental results are shown in Fig.~\ref{fig:firstMethod}.
From Fig.~\ref{fig:firstMethod}, we know that when the value of $\lambda$ is large, fewer nodes will be used, but the performance of the algorithm will be poor.
While the second method introduce a constraint to restrict its number of centers. Unlike the first method, it can directly set the maximal number of centers without introducing any regularization parameter.
With different fault scenarios, the corresponding experimental results are given by Fig.~\ref{fig:secondMethod}.
From Fig.~\ref{fig:secondMethod}, we see that if we decrease the number of nodes, the performance of the second method will be worse.

\begin{figure}[h]
\centering
%\begin{tabular}{c@{\extracolsep{10mm}}c}
%\mbox{\epsfig{figure=ASN_p_0.01_sb_0.01_MCP_convex_objective_convergence.eps,width=2in}}
\mbox{\epsfig{figure=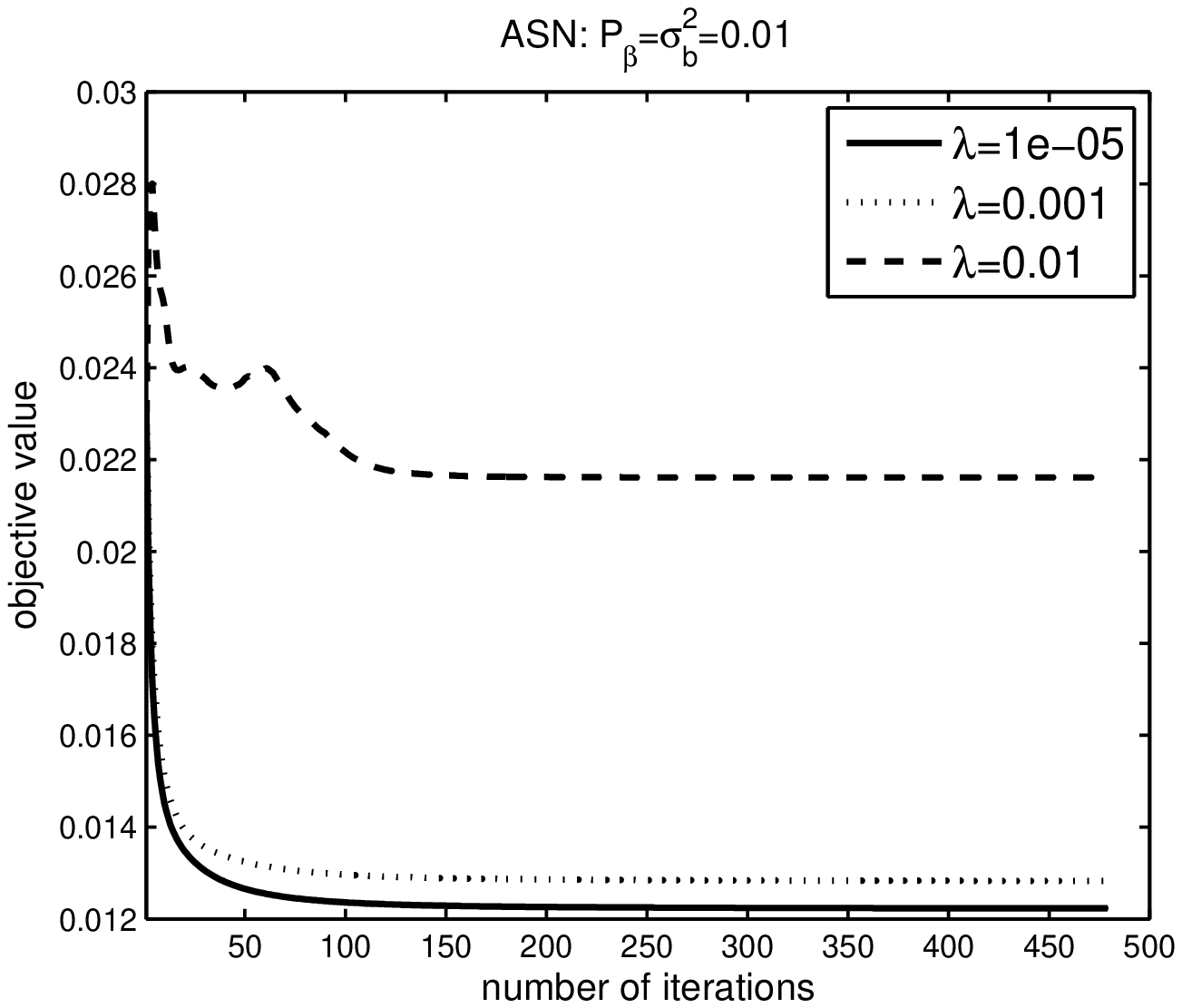,width=2in}}
\mbox{\epsfig{figure=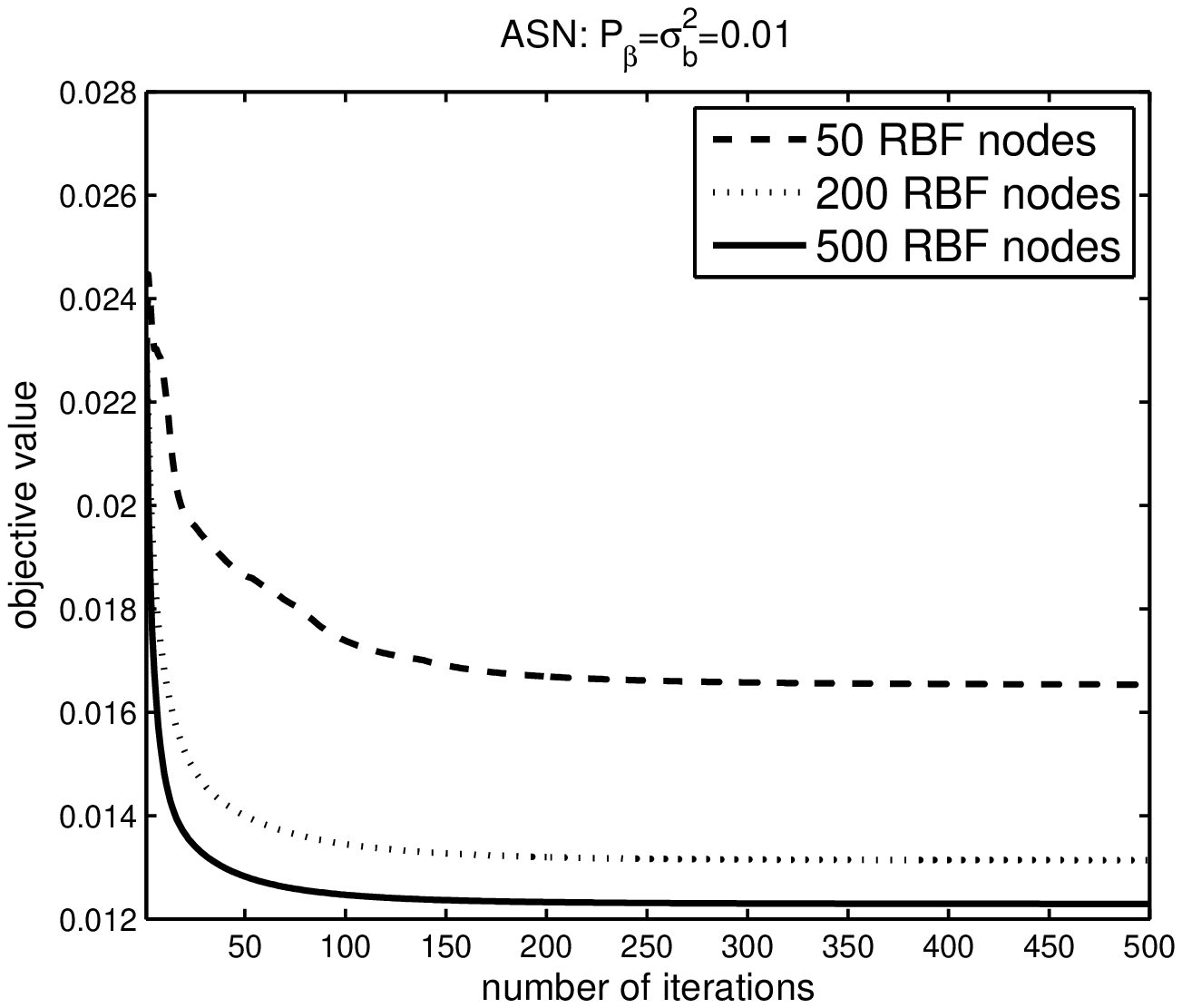,width=2in}}
%\end{tabular}
\caption{Convergence of the proposed methods. The first one is the ADMM-MCP method. The second one is the ADMM-HT.}
\label{fig:convergence}
\end{figure}
\subsection{Convergence}
The convergence of the proposed approaches has been discussed.
Here we use the dataset ASN with $\{P_\beta=\sigma_b^2=0.01\}$ as an example to intuitively demonstrate their convergence.
The results are shown in Fig.~\ref{fig:convergence}.
From Fig.~\ref{fig:convergence}, it is observed that within 100 to 200 iterations both methods converge.
If we increase the value of $\lambda$ in the first method or decrease the number of centers in the second method, these algorithms will converge to a larger objective value. For all other datasets, they have similar properties of convergence.

\begin{table}[tbp]
\newcommand{\tabincell}[2]{\begin{tabular}{@{}#1@{}}#2\end{tabular}}
\centering
\begin{tabular}{|c|c|}
\hline
Dataset &Parameters\\ \hline
ABA &\tabincell{c}{$C=\{0.01, 0.03, 0.06, 0.1, 0.3, 0.6, 1\}$, \\$\epsilon=\{1, 1.5, 2, 2.5, 3, 3.5, 4, 4.5, 5, 5.5, 6\}$}\\\hline
ASN &\tabincell{c}{$C=\{0.005, 0.01, 0.03, 0.05, 0.1, 0.3, 0.5\}$, \\$\epsilon=\{0.01, 0.05, 0.1, 0.15, 0.175, 0.2, 0.25, 0.3, 0.35, 0.4\}$}\\\hline
HOUSING &\tabincell{c}{$C=\{0.01, 0.02, 0.04, 0.08, 0.2, 0.4, 0.8\}$, \\$\epsilon=\{0.01, 0.02, 0.04, 0.08, 0.2, 0.4, 0.8\}$}\\\hline
CON &\tabincell{c}{$C=\{0.01, 0.03, 0.06, 0.1, 0.3, 0.6, 1\}$, \\ $\epsilon=\{0.01, 0.03, 0.06, 0.1, 0.3, 0.6, 1\}$}\\\hline
ENERGY &\tabincell{c}{$C=\{0.005, 0.01, 0.05, 0.1, 0.3, 0.5\}$, \\$\epsilon=\{0.01, 0.05, 0.1, 0.125, 0.15, 0.2, 0.25, 0.3, 0.35, 0.4\}$}\\\hline
WQW &\tabincell{c}{$C=\{0.001, 0.005, 0.01, 0.05, 0.1,0.2\}$, \\$\epsilon=\{0.005, 0.0075, 0.01, 0.025, 0.05, 0.075, 0.1, 0.2\}$}\\ \hline
\end{tabular}
\caption{Tuning parameter settings of SVR algorithm.}
\label{SVRparam}
\end{table}

\begin{figure*}[h]
\begin{tabular}{c@{\extracolsep{2mm}}c@{\extracolsep{2mm}}c}
\mbox{\epsfig{figure=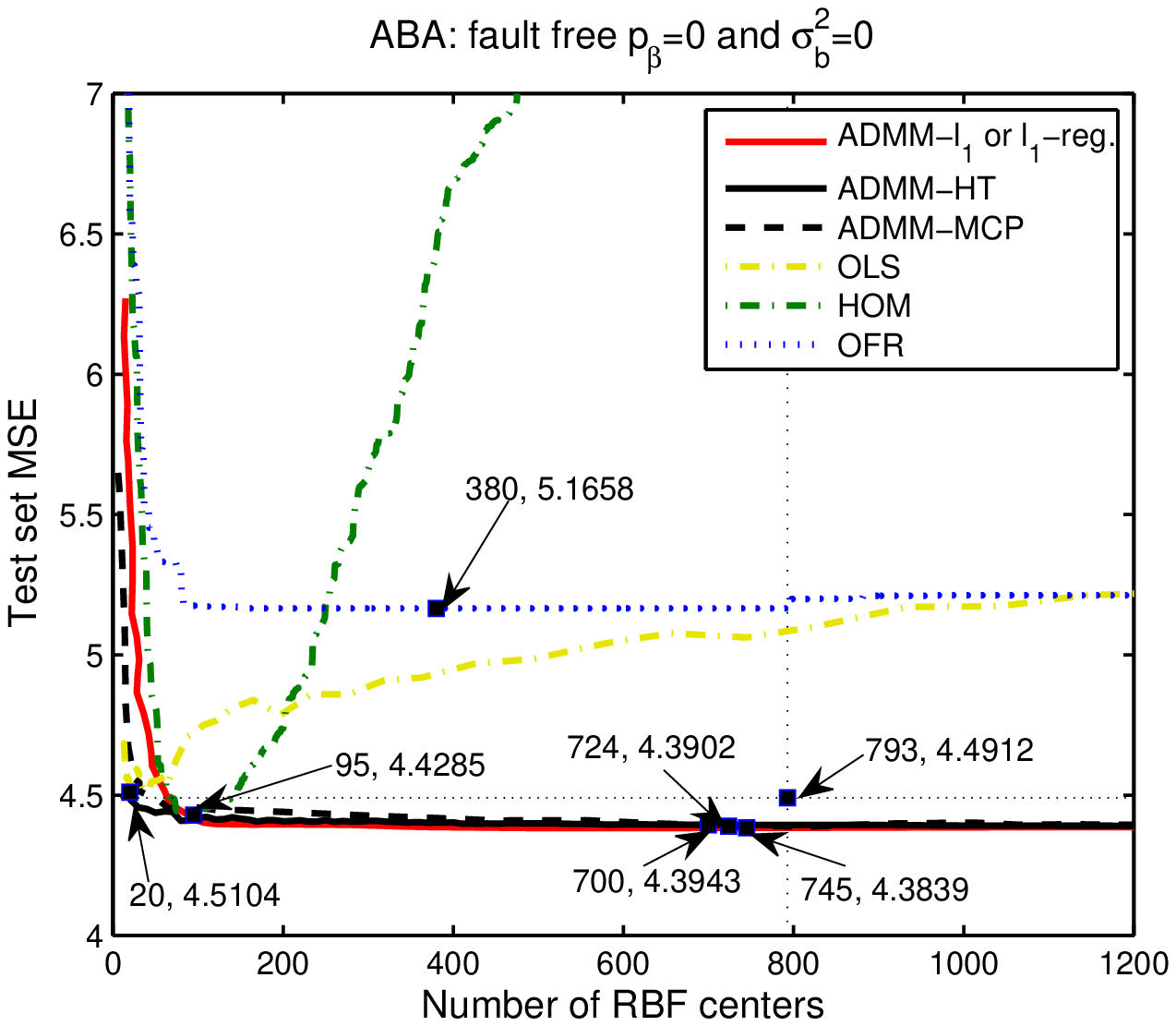,width=2.2in}} &
\mbox{\epsfig{figure=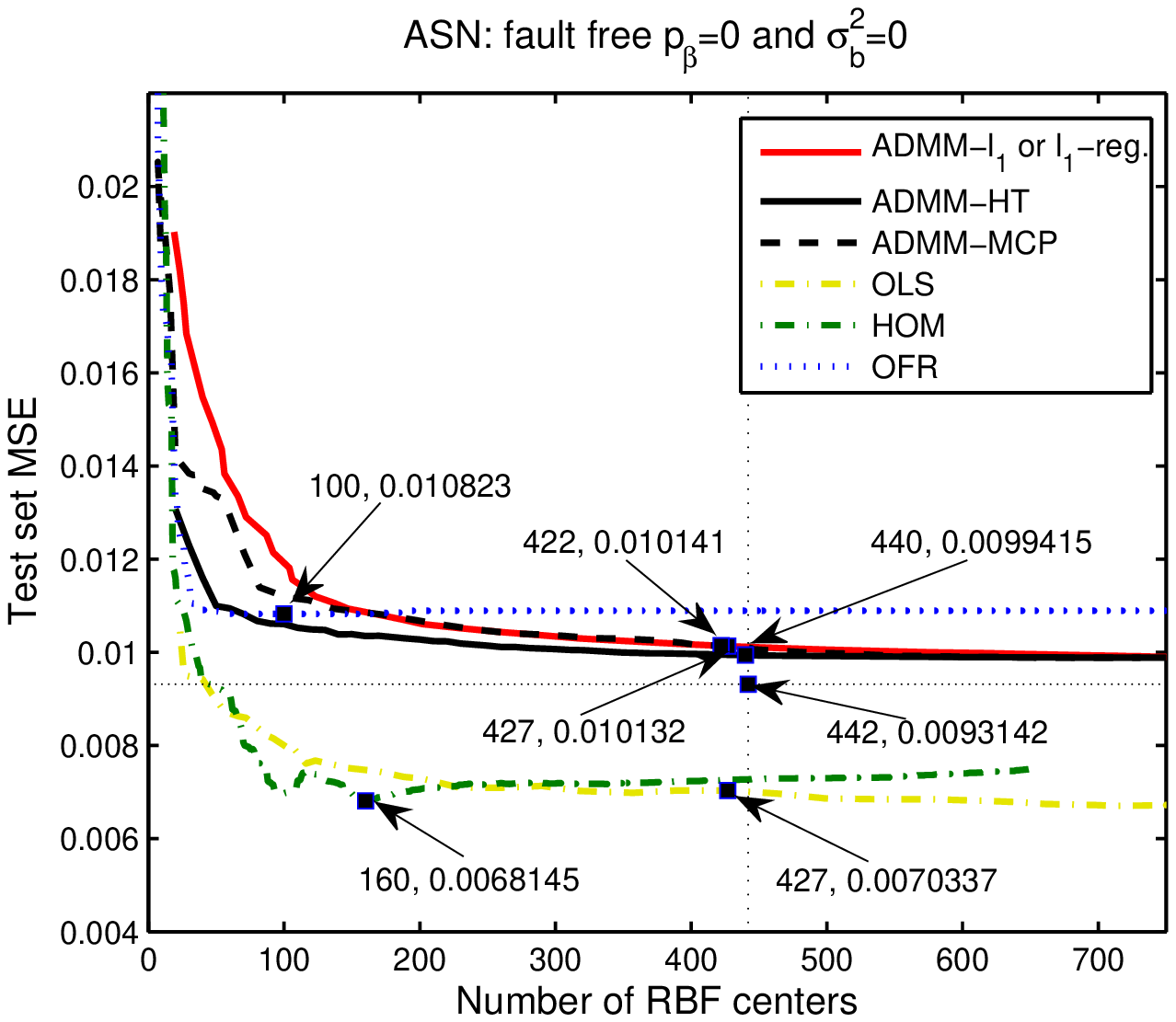,width=2.2in}} &
\mbox{\epsfig{figure=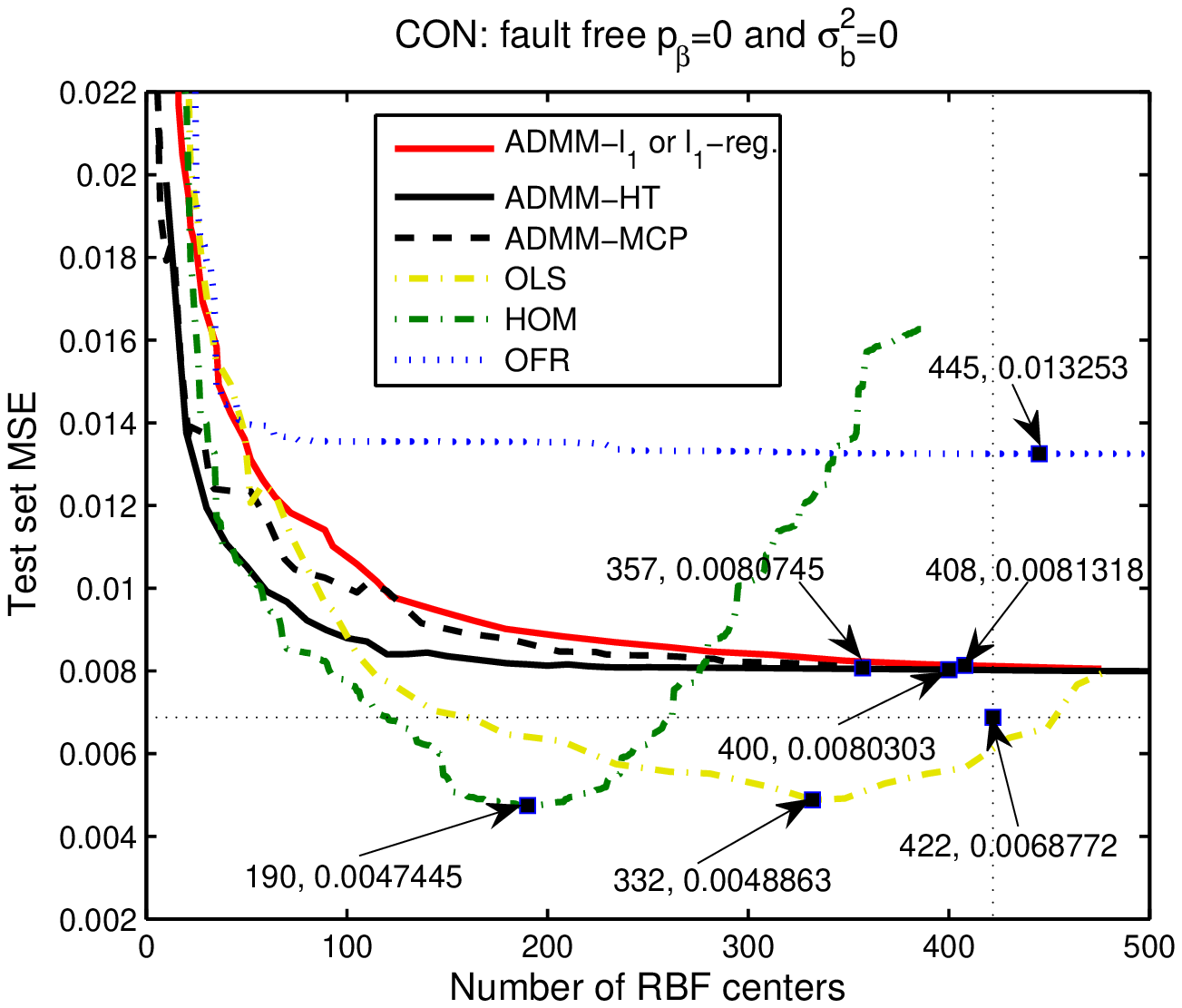,width=2.2in}}\\
\mbox{\epsfig{figure=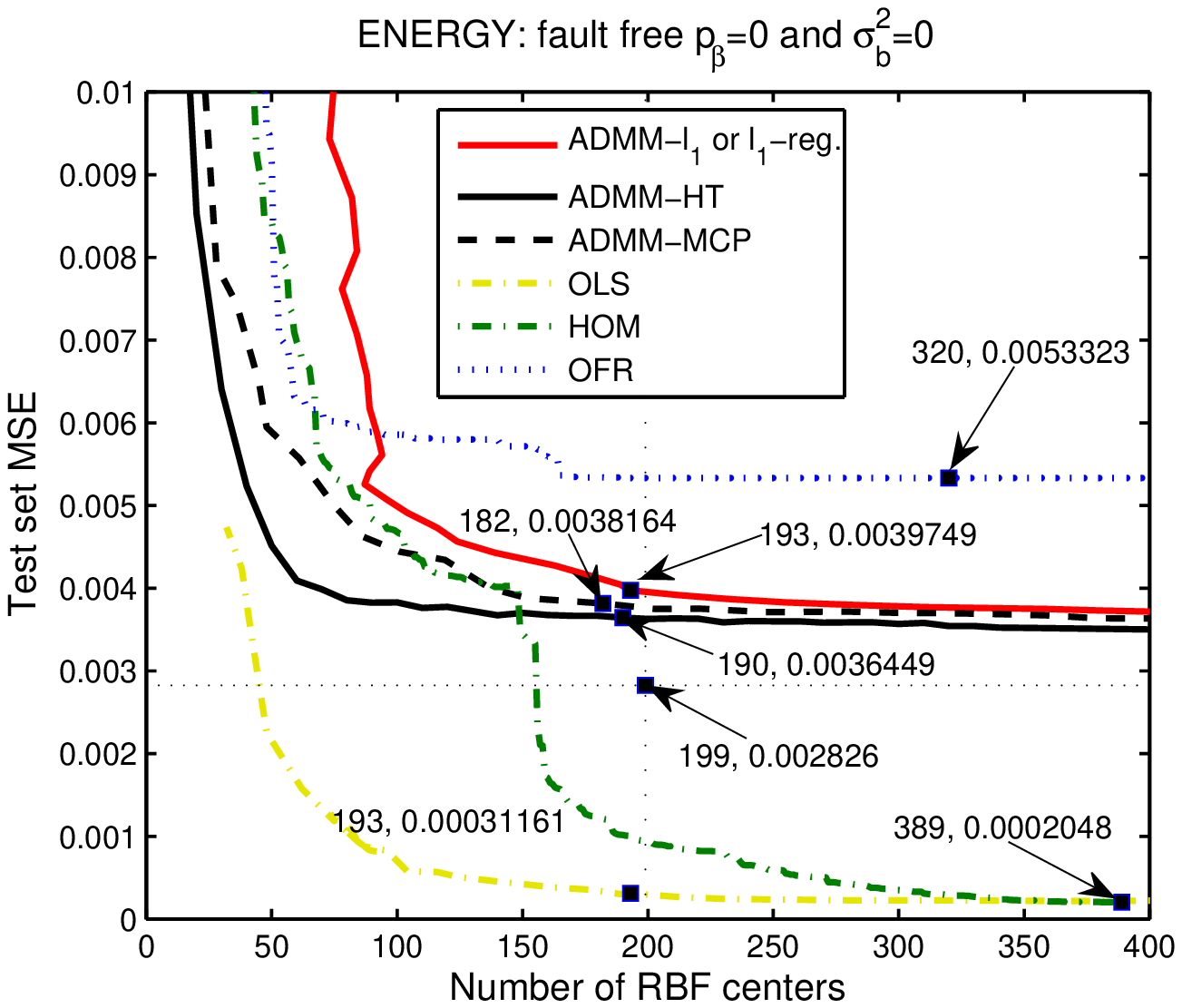,width=2.2in}} &
\mbox{\epsfig{figure=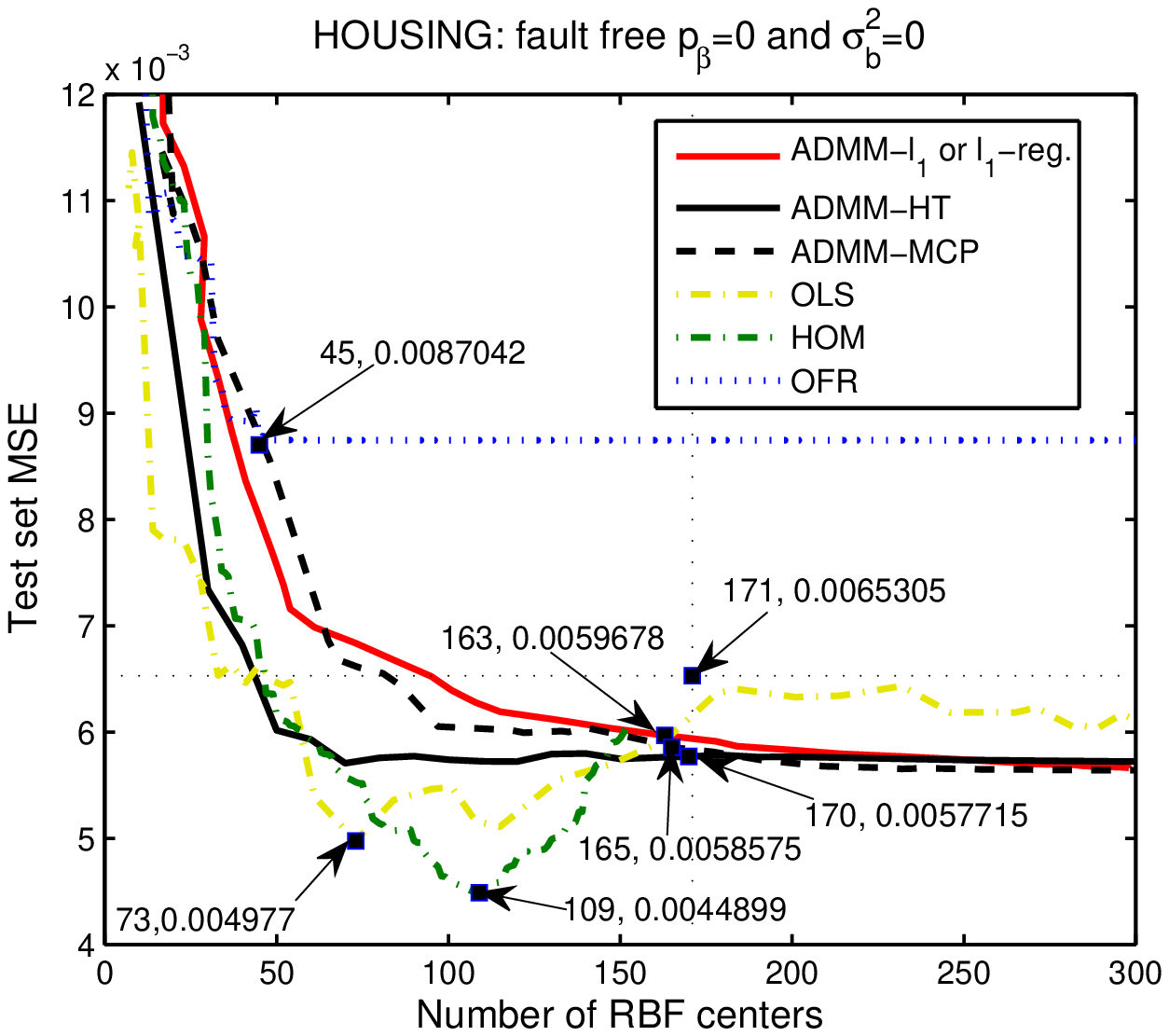,width=2.2in}} &
\mbox{\epsfig{figure=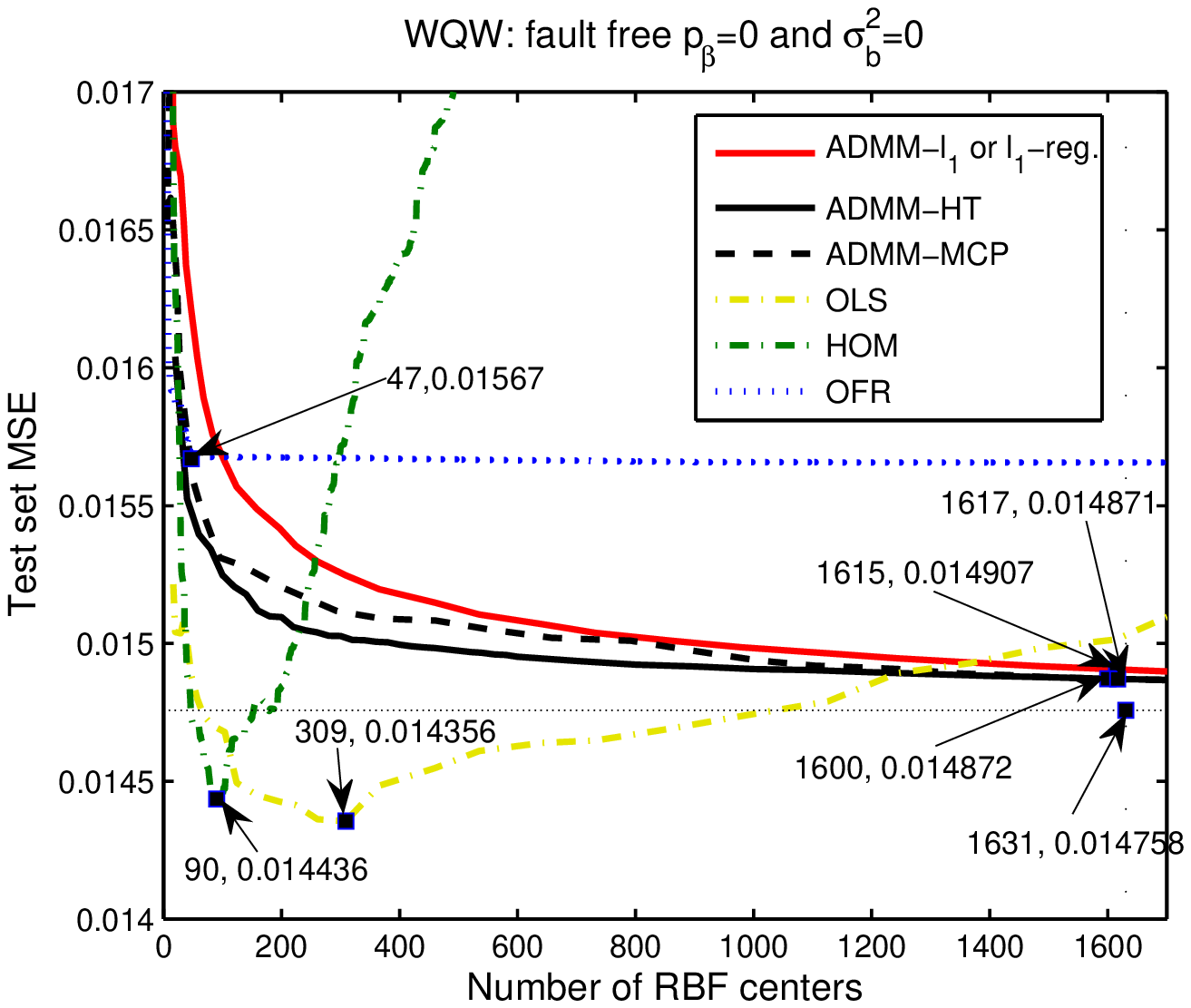,width=2.2in}}
\end{tabular}
\caption{Performance of different algorithms under fault-free situation.}
\label{fig:faultfree}
\end{figure*}

\begin{table*}[tbp]
\newcommand{\tabincell}[2]{\begin{tabular}{@{}#1@{}}#2\end{tabular}}
\centering
\renewcommand{\multirowsetup}{\centering}
\begin{tabular}{|c|c|c|c|c|c|c|c|c|}
\hline
\multirow{2}{*}{Dataset} &\multicolumn{2}{c|}{ADMM-HT}
&\multicolumn{2}{c|}{ADMM-MCP} %&\multicolumn{2}{c|}{convex ADMM-MCP}
&\multicolumn{2}{c|}{ADMM-$l_1$} &\multicolumn{2}{c|}{OLS}\\
\cline{2-9}
& \tabincell{c}{AVG \\MSE} & \tabincell{c}{AVG no.\\of nodes}
& \tabincell{c}{AVG \\MSE} & \tabincell{c}{AVG no.\\of nodes}
%& \tabincell{c}{AVG \\MSE} & \tabincell{c}{AVG no.\\of nodes}
& \tabincell{c}{AVG \\MSE} & \tabincell{c}{AVG no.\\of nodes}
& \tabincell{c}{AVG \\MSE} & \tabincell{c}{AVG no.\\of nodes}\\
\hline
ABA &4.579 &730.0 &4.579 &730.2  &4.570 &726.4 &4.641 &40.8\\
\hline
ASN &0.01076 &412.0&0.01100 &409.4 &0.01096 &401.0 &0.00736 &401.0\\
\hline
HOUSING &0.00688 &136.5 &0.00745 &135.3  &0.00746 &134.2 &0.00688 &129.7\\
\hline
CON &0.00839 &337.0 &0.00848 &327.0 &0.0086 &351.6 &0.0066 &203.4\\
\hline
ENERGY &0.00452 &325.5 &0.00453 &328.5 &0.00459 &324.5 &0.00036 &324.5\\
\hline
WQW &0.01471 &1475.0 &0.01473 &1490.0 &0.01476 &1460.0 & 0.01424 &338.0\\
\hline
\end{tabular}
\begin{tabular}{|c|c|c|c|c|c|c|c|c|}
\hline
\multirow{2}{*}{Dataset}
&\multicolumn{2}{c|}{$l_1$-reg.} &\multicolumn{2}{c|}{SVR}
&\multicolumn{2}{c|}{HOM} &\multicolumn{2}{c|}{OFR}
\\
\cline{2-9}
& \tabincell{c}{AVG \\MSE} & \tabincell{c}{AVG no.\\of nodes}
& \tabincell{c}{AVG \\MSE} & \tabincell{c}{AVG no.\\of nodes}
& \tabincell{c}{AVG \\MSE} & \tabincell{c}{AVG no.\\of nodes}
& \tabincell{c}{AVG \\MSE} & \tabincell{c}{AVG no.\\of nodes}\\
\hline
ABA &4.57 &726.4 &4.749	&777.4	&4.592 &95.6 &5.228	&673.7\\
\hline
ASN &0.01096 &401.0 &0.01020 &418.1 &0.00667 &200.2 &0.01275 &319.1\\
\hline
HOUSING &0.00746 &134.2 &0.00782 &141.4 &0.00567 &104.9 &0.01248 &55.1\\
\hline
CON &0.00860 &351.6 &0.00719 &364.5 &0.00632 &179.6 &0.01329 &250.7\\
\hline
ENERGY &0.00459 &324.5 &0.00340 &339.7 &0.00290 &380.3 &0.00549 & 190.2\\
\hline
WQW &0.01480 &1468.0 &0.01471 &1514.3 &0.01417 &146.8 &0.01505 &563.5\\
\hline
\end{tabular}
\caption{Average test MSE over 20 trials under fault-free situation.}
\label{resultFaultFree}
\end{table*}

\begin{figure*}[h]
\begin{tabular}{c@{\extracolsep{2mm}}c@{\extracolsep{2mm}}c}
\mbox{\epsfig{figure=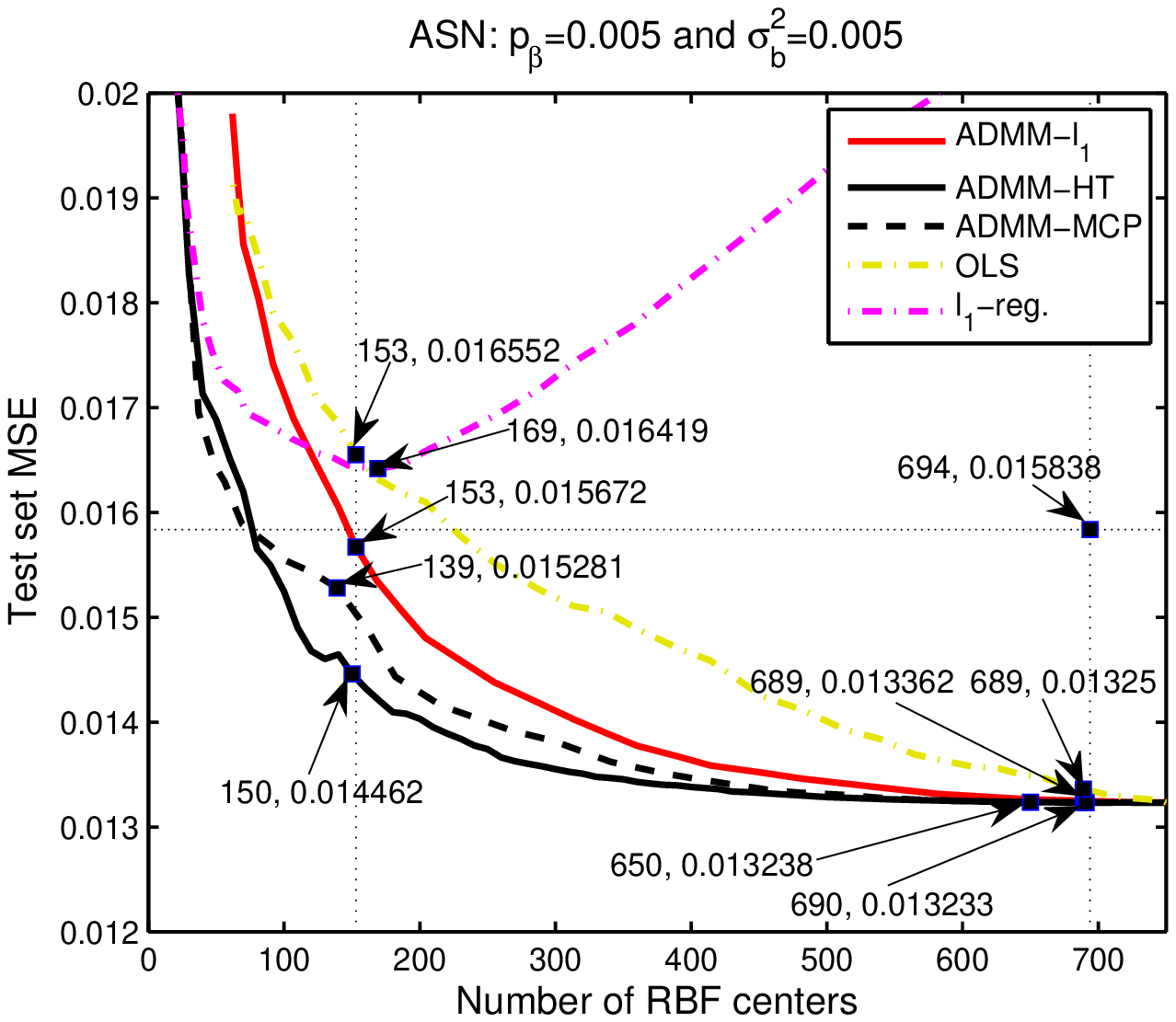,width=2.2in}} &
\mbox{\epsfig{figure=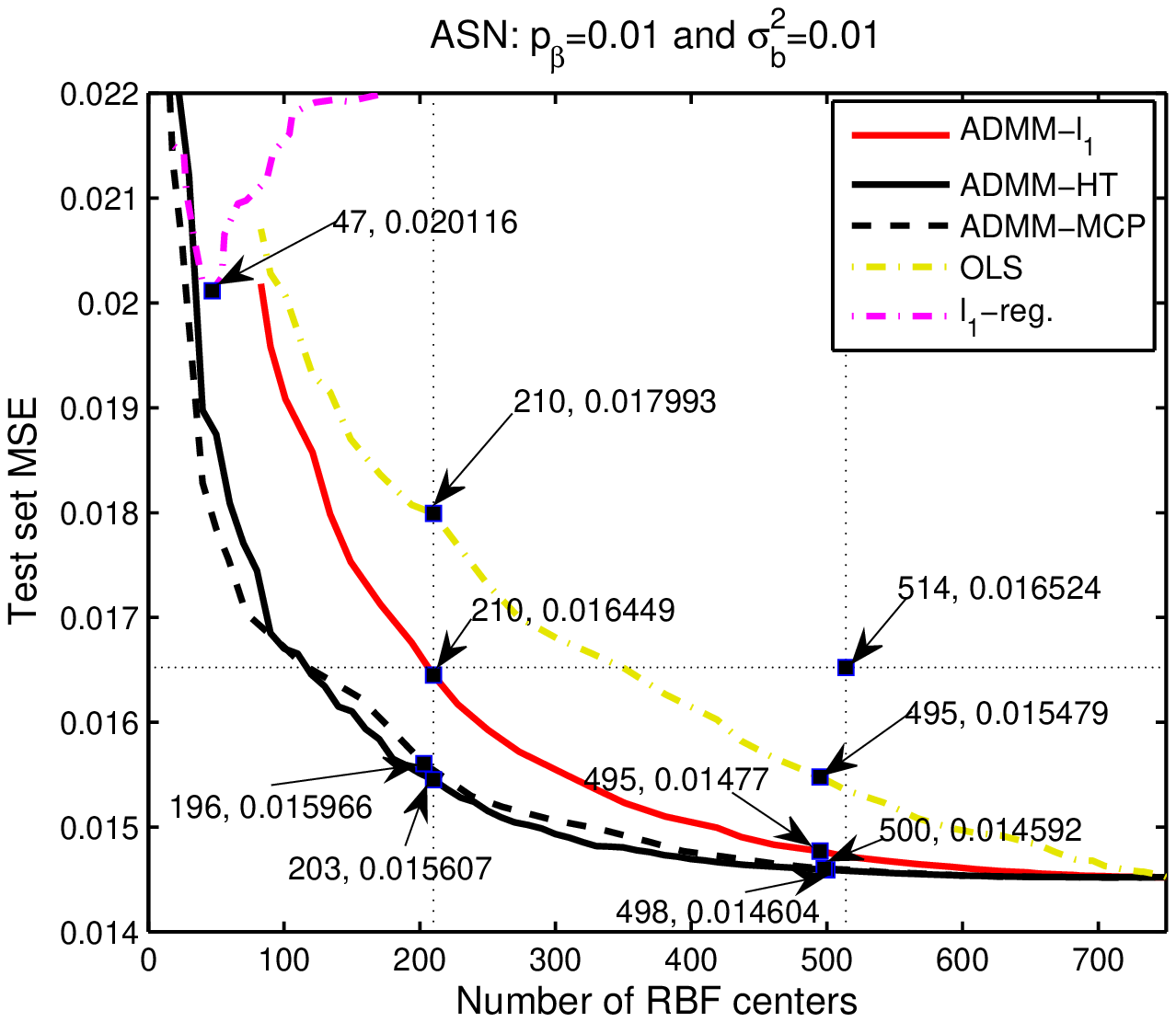,width=2.2in}} &
\mbox{\epsfig{figure=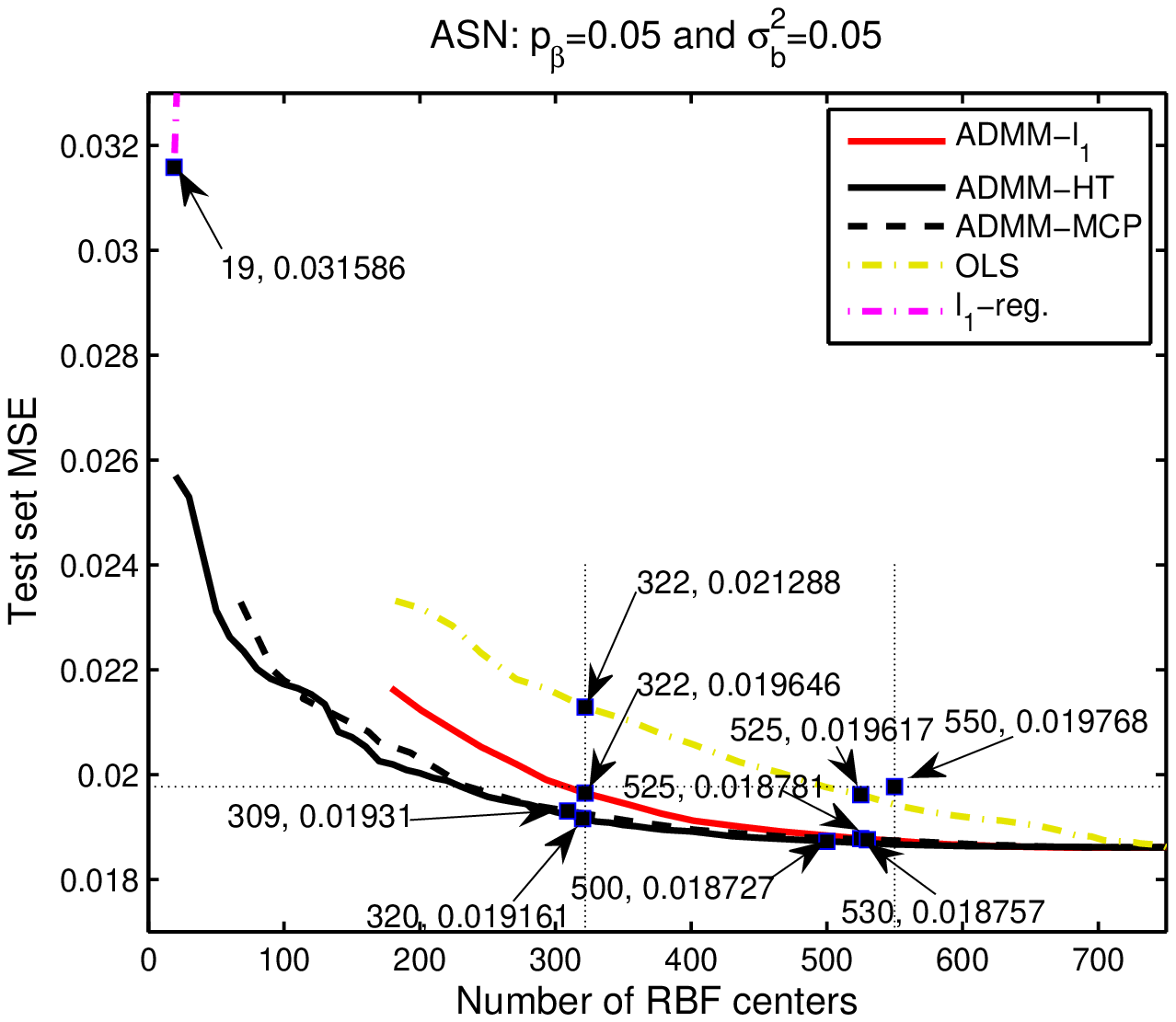,width=2.2in}}\\
\mbox{\epsfig{figure=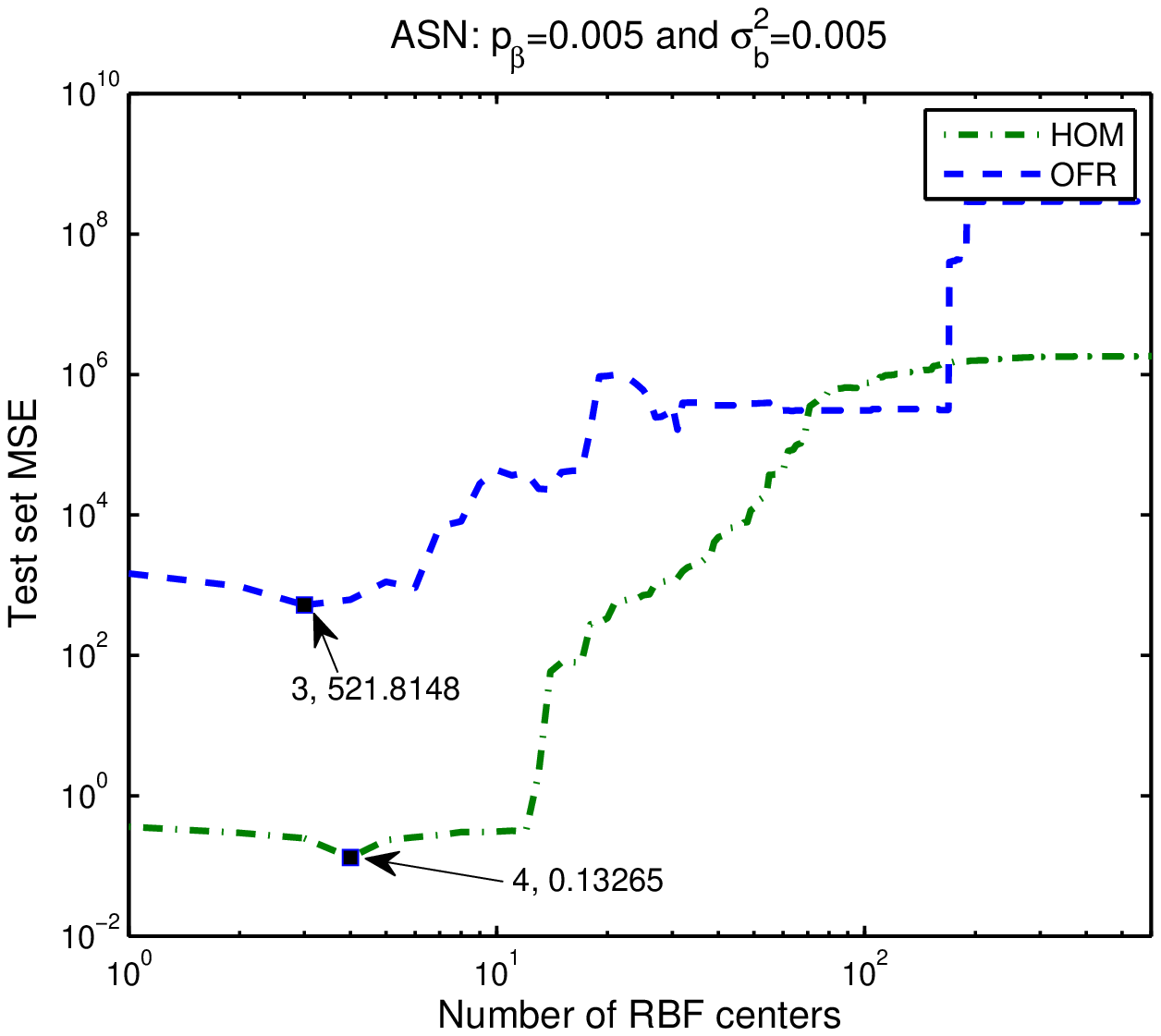,width=2.2in}} &
\mbox{\epsfig{figure=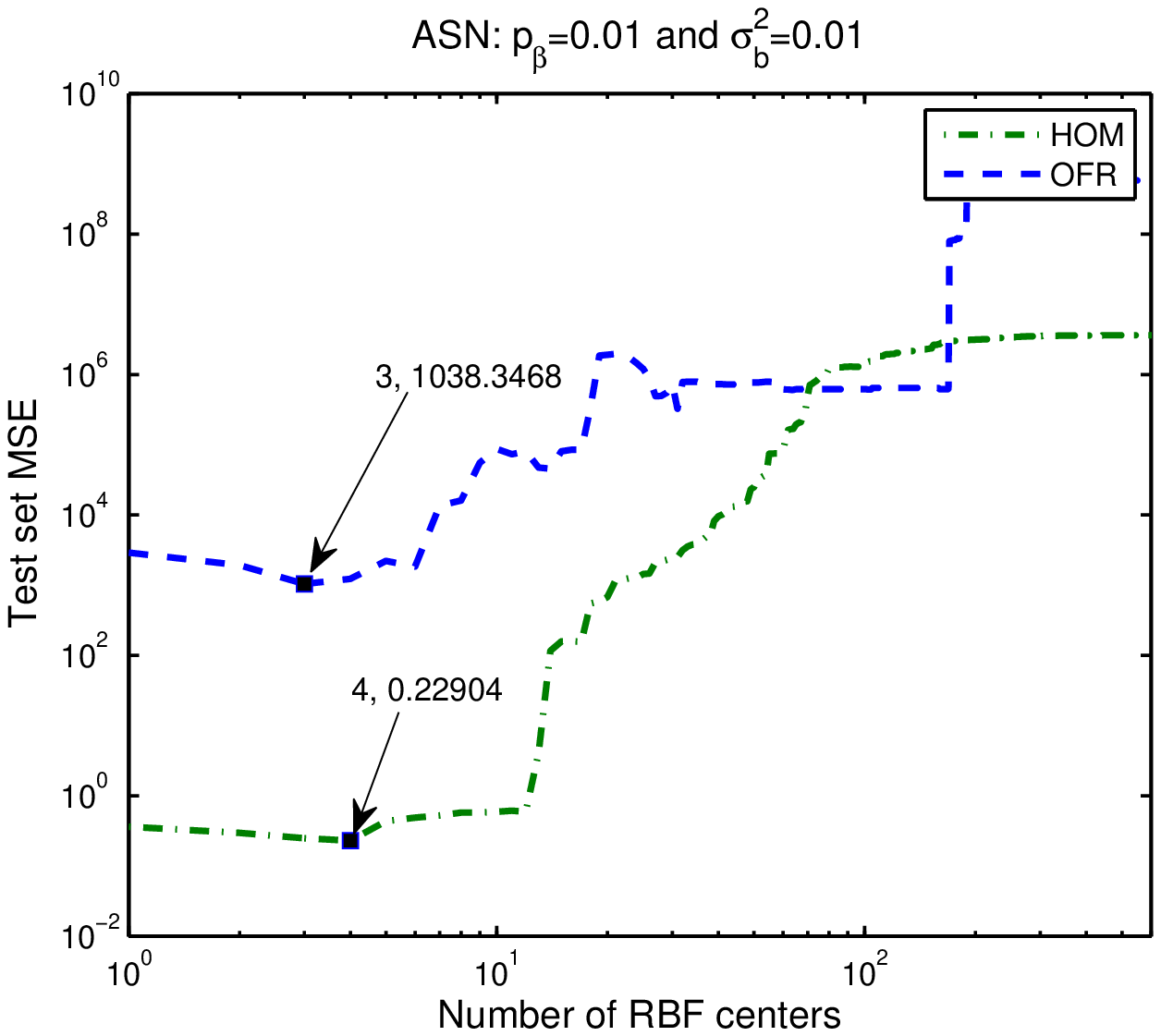,width=2.2in}} &
\mbox{\epsfig{figure=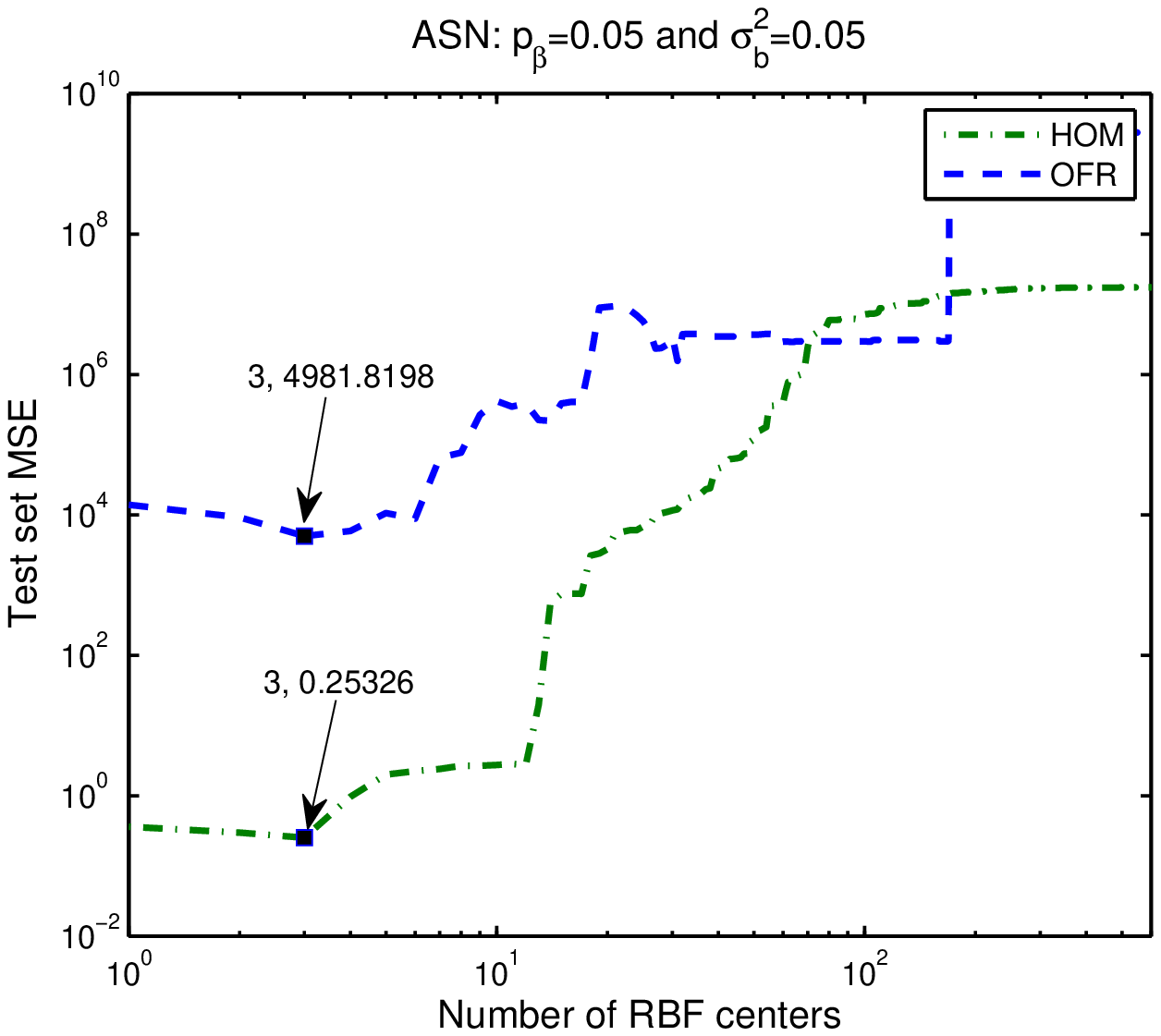,width=2.2in}}
\end{tabular}
\caption{Performance of different algorithms under concurrent fault situation.}
\label{fig:fault}
\end{figure*}

\begin{table*}[tbp]
\newcommand{\tabincell}[2]{\begin{tabular}{@{}#1@{}}#2\end{tabular}}
\centering
\renewcommand{\multirowsetup}{\centering}
\begin{tabular}{|c|c|c|c|c|c|c|c|c|c|}
\hline
\multirow{2}{*}{Dataset} &\multirow{2}{*}{Fault level} &\multicolumn{2}{c|}{ADMM-HT}
&\multicolumn{2}{c|}{ADMM-MCP} %&\multicolumn{2}{c|}{convex ADMM-MCP}
&\multicolumn{2}{c|}{ADMM-$l_1$} &\multicolumn{2}{c|}{OLS}\\
\cline{3-10}
& & \tabincell{c}{AVG \\MSE} & \tabincell{c}{AVG no.\\of nodes}
%& \tabincell{c}{AVG \\MSE} & \tabincell{c}{AVG no.\\of nodes}
& \tabincell{c}{AVG \\MSE} & \tabincell{c}{AVG no.\\of nodes}
& \tabincell{c}{AVG \\MSE} & \tabincell{c}{AVG no.\\of nodes}
& \tabincell{c}{AVG \\MSE} & \tabincell{c}{AVG no.\\of nodes}\\
\hline
\multirow{3}{*}{ABA} &$P_\beta=\sigma_b^2=0.005$ &5.1172 &152.0 &5.1617 &140.9
&5.3611 &156.2 &5.3269 &156.2\\ \cline{2-10}
&$P_\beta=\sigma_b^2=0.01$ &5.223 &198.0 &5.2587 &174.8  &5.5254 &202.2 &5.4416 &202.2 \\ \cline{2-10}
&$P_\beta=\sigma_b^2=0.05$ &5.6029 &353.5 &5.6169 &313.8  &6.2372 &356.8 &5.8697 &356.8 \\ \hline
\multirow{3}{*}{ASN} &$P_\beta=\sigma_b^2=0.005$ &0.01490 &150.0 &0.01512 &142.4 &0.01617 &154.6 &0.01810 &154.6 \\ \cline{2-10}
&$P_\beta=\sigma_b^2=0.01$ &0.01597 &204.5 &0.01596 &193.9  &0.01711 &209.1 &0.01928 &209.1\\ \cline{2-10}
&$P_\beta=\sigma_b^2=0.05$ &0.01992 &341.0 &0.02006 &323.4 &0.02086 &347.2 &0.02262 &347.2\\ \hline
\multirow{3}{*}{HOUSING} &$P_\beta=\sigma_b^2=0.005$ &0.01511 &57.5 &0.01527 &56.8 &0.01583 &61.9 &0.01654 &61.9\\ \cline{2-10}
&$P_\beta=\sigma_b^2=0.01$ &0.01643 &56.5 &0.01656 &53.6 &0.01743 &60.9 &0.01832 &60.9\\ \cline{2-10}
&$P_\beta=\sigma_b^2=0.05$ &0.02129 &57.5 &0.02135 &55.9 &0.02343 &62.0 &0.02476 &62.0 \\ \hline
\multirow{3}{*}{CON} &$P_\beta=\sigma_b^2=0.005$ &0.01215 &127.5 &0.01218 &124.2  &0.01334 &131.5 &0.01756 &131.5 \\ \cline{2-10}
&$P_\beta=\sigma_b^2=0.01$ &0.01352 &118.5 &0.01387 &114.7  &0.01489 &122.6 &0.01978 &122.6 \\ \cline{2-10}
&$P_\beta=\sigma_b^2=0.05$ &0.01756 &135.0 &0.01724 &133.1  &0.01910 &140.0 &0.02515 &140.0 \\ \hline
\multirow{3}{*}{ENERGY} &$P_\beta=\sigma_b^2=0.005$ &0.00519 &157.0 &0.00518 &150.2 &0.00560 &161.1 &0.00568 &161.1 \\ \cline{2-10}
&$P_\beta=\sigma_b^2=0.01$ &0.00563 &162.5 &0.00560 &156.1  &0.00610 &167.0 &0.00632 &167.0 \\ \cline{2-10}
&$P_\beta=\sigma_b^2=0.05$ &0.00761 &229.0 &0.00761 &222.2  &0.00856 &233.3 &0.00849 &233.3 \\ \hline
\multirow{3}{*}{WQW} &$P_\beta=\sigma_b^2=0.005$ &0.01643 &110.0 &0.01641 &106.6 &0.01678 &119.6 &0.01732 &119.6\\ \cline{2-10}
&$P_\beta=\sigma_b^2=0.01$ &0.01665 &160.0 &0.01660 &142.6 &0.01701 &169.2 &0.01753 &169.2 \\ \cline{2-10}
&$P_\beta=\sigma_b^2=0.05$ &0.01730 &373.0 &0.01729 &355.8 &0.01763 &385.6 &0.01807 &385.6\\ \hline
\end{tabular}
\begin{tabular}{|c|c|c|c|c|c|c|c|c|c|}
\hline
\multirow{2}{*}{Dataset} &\multirow{2}{*}{Fault level}
&\multicolumn{2}{c|}{$l_1$-reg.} &\multicolumn{2}{c|}{SVR}
&\multicolumn{2}{c|}{HOM} &\multicolumn{2}{c|}{OFR}
\\
\cline{3-10}
& & \tabincell{c}{AVG \\MSE} & \tabincell{c}{AVG no.\\of nodes}
& \tabincell{c}{AVG \\MSE} & \tabincell{c}{AVG no.\\of nodes}
& \tabincell{c}{AVG \\MSE} & \tabincell{c}{AVG no.\\of nodes}
& \tabincell{c}{AVG \\MSE} & \tabincell{c}{AVG no.\\of nodes}\\
\hline
\multirow{3}{*}{ABA} &$P_\beta=\sigma_b^2=0.005$ &5.5938 &28.3 &5.388 &767 &51.3834 &7.5  &580584	&3.1 \\ \cline{2-10}
&$P_\beta=\sigma_b^2=0.01$ &5.9977 &25.1 &5.5537 &748.0 &56.2897 &5.4 &1155290 &3.1\\ \cline{2-10}
&$P_\beta=\sigma_b^2=0.05$ &8.4599 &19.2 &6.2801 &854.6 &62.6329 &3.9 &5542897 &3.1\\ \hline
\multirow{3}{*}{ASN} &$P_\beta=\sigma_b^2=0.005$ &0.01672 &165.2 &0.01641 &467.2 &0.08730 &3.3 &569.2 &2.8 \\ \cline{2-10}
&$P_\beta=\sigma_b^2=0.01$ &0.02054 &57.2 &0.01731 &609.3 &0.09517 &3.1 &1132.7 &2.8 \\ \cline{2-10}
&$P_\beta=\sigma_b^2=0.05$ &0.03078 &20.0 &0.02105 &621.4 &0.11130 &3.0 &5434.4 &2.8 \\ \hline
\multirow{3}{*}{HOUSING} &$P_\beta=\sigma_b^2=0.005$ &0.01798 &12.9 &0.01716 &88.2 &0.05489 &4.7 &3.721 &2.4 \\ \cline{2-10}
&$P_\beta=\sigma_b^2=0.01$ &0.02053 &10.6 &0.01883 &75.5 &0.06362 &4.0 &7.362 &2.4\\ \cline{2-10}
&$P_\beta=\sigma_b^2=0.05$ &0.02999 &6.9 &0.02476 &87.9 &0.07864 &3.4 &35.160 &2.4\\ \hline
\multirow{3}{*}{CON} &$P_\beta=\sigma_b^2=0.005$ &0.01258 &138.5 &0.01376 &278.5 &0.04479 &13.1 &89.61 &4.1 \\ \cline{2-10}
&$P_\beta=\sigma_b^2=0.01$ &0.01483 &94.1 &0.01514 &271.8 &0.05112 &11.3 &178.25 &4.1\\ \cline{2-10}
&$P_\beta=\sigma_b^2=0.05$ &0.02358 &32.3 &0.01942 &284.9 &0.06605 &8.9 &855.10 &4.1\\ \hline
\multirow{3}{*}{ENERGY} &$P_\beta=\sigma_b^2=0.005$ &0.00542 &155.0 &0.00566 &293.1 &0.02966 &26.5 &0.05854 &9.3\\ \cline{2-10}
&$P_\beta=\sigma_b^2=0.01$ &0.00591 &159.0 &0.00619 &363.8 &0.03859 &21.8 &0.06895 &8.4 \\ \cline{2-10}
&$P_\beta=\sigma_b^2=0.05$ &0.01026 &123.4 &0.00870 &211.1 &0.06420 &14.3 &0.10650 &4.8 \\ \hline
\multirow{3}{*}{WQW} &$P_\beta=\sigma_b^2=0.005$ &0.01702 &39.6 &0.01685 &1346.1 &0.03350 &5.2 &535.7 &2.2 \\ \cline{2-10}
&$P_\beta=\sigma_b^2=0.01$ &0.01788 &29.6 &0.01710 &1461.0 &0.03970 &5.0 &1065.9 &2.2 \\ \cline{2-10}
&$P_\beta=\sigma_b^2=0.05$ &0.02419 &22.6 &0.01811 &697.8 &0.06621 &4.4 &5114.1 &2.2\\ \hline
\end{tabular}
\caption{Average test MSE over 20 trials under concurrent faults.}
\label{resultFault}
\end{table*}

\subsection{Comparison}
In this subsection, we compare our two methods with six other algorithms. They are, respectively, the fault tolerant OLS algorithm (OLS) \cite{leung2017regularizer}, the fault tolerant $l_1$-norm approach (ADMM-$l_1$) \cite{wang2017admm}, the $l_1$-norm regularization approach ($l_1$-reg.) \cite{zhang2015comparison}, the support vector regression algorithm (SVR) \cite{zhang2015comparison}, the orthogonal forward regression algorithm (OFR) \cite{hong2015nonlinear} and the Homotopy method (HOM) \cite{malioutov2005homotopy}. %The fault tolerant OLS algorithm and the fault tolerant $l_1$-norm algorithm are two fault tolerant method. %As far as we know the performance of them are better than other existing fault tolerant methods including the optimized weight decay method \cite{paper:leung2012}, the replication method \cite{Mahiani2012,graham2004nanocomputing} and Zhou's method \cite{paper:zhou03}.

The fault tolerant OLS algorithm includes two stages. In the first one, it uses OLS method to generate a sorted list of RBF nodes. In the second stage, it constructs a fault tolerant RBF network with desired number of nodes.
The fault tolerant $l_1$-norm approach is our previous work. It selects centers and constructs the fault tolerant RBF network simultaneously. But compared with the fault tolerant OLS algorithm, the improvement of the fault tolerant $l_1$-norm approach is not very significant.

The $l_1$-norm regularization approach \cite{zhang2015comparison} also uses the $l_1$-norm to control the nodes' number used in the RBF network. But its fault tolerant ability is inadequate. Especially, when the fault level is high.

The SVR algorithm \cite{zhang2015comparison} can also train the RBF network and select centers simultaneously.
The parameters $C$ and $\epsilon$ are used to control the training process. TABLE~\ref{SVRparam} shows the parameter settings for different datasets.
The SVR algorithm has fault tolerant ability.
Since the parameter $C$ is capable to limit the magnitudes of the trained weights.
The parameter $\epsilon$ is used to control its approximation ability.
However, the main drawback of the SVR algorithm is that there is no simple way to find an appropriate pair of $C$ and $\epsilon$.
In our experiment, we use trial-and-error method to determine them.

The Homotopy method \cite{malioutov2005homotopy} is an incremental learning method.
It also uses an $l_1$-norm regularization term, and it can tune the regularization parameter automatically.
The OFR algorithm \cite{hong2015nonlinear} is also an incremental learning method.
It chooses one RBF center at a time with the orthogonal forward regression procedure.
For OFR, an $l_2$-norm regularization term is used.
And it can also tune the regularization parameter automatically during training process.

In the following two experiments, the simulation are ran 20 times. And in each trial, the samples of dataset are randomly assigned to the training set and testing set.
First, we compare the proposed methods with all above mentioned approaches under the fault-free situation. The typical examples are given by Fig.~\ref{fig:faultfree}. In this case, the performance of the fault tolerant $l_1$-norm approach and the $l_1$-norm regularization approach are substantially same with each other.
For OLS, HOM, OFR and SVR, we select their minimum MSE and the corresponding number of nodes to represent their performance.
For other methods, we use the points where the number of nodes is close to the best result of SVR to represent their performance.
It is because that their MSEs are basically decreasing with the increasing of nodes' number.
%And the minimum MSEs of SVR algorithm are given by the independent points in Fig.~\ref{fig:faultfree}.

TABLE~\ref{resultFaultFree} shows the average test set error over the 20 trials. From this table and Fig.~\ref{fig:faultfree}, it is observed that, under fault-free environment, the performance of OLS and HOM are better than others.

Next, we compare the proposed methods with all other algorithms under concurrent faults.
We use three different fault levels: $\{P_\beta=\sigma_b^2=0.005\}$, $\{P_\beta=\sigma_b^2=0.01\}$ and $\{P_\beta=\sigma_b^2=0.05\}$.
The typical results of one trial for ASN dataset under different fault levels are given by Fig.~\ref{fig:fault}.
Where the first column is the results when $\{P_\beta=\sigma_b^2=0.005\}$, while the second column and the third one are respectively the results when $\{P_\beta=\sigma_b^2=0.01\}$ and $\{P_\beta=\sigma_b^2=0.05\}$.
Then, we use the first column as an example to explain these figures.
The independent point $(694,0.001538)$ in the first figure is the best result of SVR algorithm.
When we use the similar number of nodes, the results of ADMM-HT, ADMM-MCP, convex ADMM-MCP, ADMM-$l_1$ and OLS are all similar with each other but better than the SVR method.
However, when using fewer centers, such as $150$, the performance of the proposed algorithms outperforms others.
The second figure in the first column shows the results of HOM and OFR.
Both of them break down under concurrent faults.
Their minimum test set MSEs are marked in the figure.
For other trials and datasets, the results are similar.

%From the figure, for each algorithm, we also select one point to represent its performance.
%For instance, when $\{P_\beta=\sigma_b^2=0.005\}$, we select the best results of SVR, $l_1$-reg., HOM, and OFR.
In each trial, the best results of SVR, $l_1$-reg., HOM, and OFR are selected to represent their performance.
For all other algorithms, their MSEs are basically always decreasing with the increasing number of nodes.
For ADMM-$l_1$, we choose the point with similar performance to the best result of SVR to represent its performance.
For ADMM-HT, ADMM-MCP and OLS, the points with similar number of nodes to the selected point of ADMM-$l_1$ are used.
After 20 times trials, we calculate the average test set error and average number of nodes for each algorithm.
The results are shown in TABLE~\ref{resultFault}.
From this table, we see that, under the concurrent fault situation, even the best results of SVR, $l_1$-reg., HOM, and OFR are used, the performance of them is still unacceptable. %Especially, when the fault level is high.
However, the ADMM-HT, ADMM-MCP, ADMM-$l_1$ and OLS can effectively reduce the influence of the concurrent faults.
Among them, the ADMM-HT and ADMM-MCP are always the best which have smaller average MSE and use fewer number of nodes.
Comparing the proposed two methods, if we carefully tune parameters $\lambda$ and $\gamma$ in ADMM-MCP method, it may have better performance than ADMM-HT.
But the most attractive thing of ADMM-HT is that we can directly select the number of nodes without tuning any indirect parameter. %even we also use the approximate $l_0$-norm in this method.

\begin{table*}[tbp]
\newcommand{\tabincell}[2]{\begin{tabular}{@{}#1@{}}#2\end{tabular}}
\centering
\renewcommand{\multirowsetup}{\centering}
\begin{tabular}{|c|c|c|c|c|c|c|c|c|c|c|c|}
\hline
\multirow{2}{*}{Dataset} &\multirow{2}{*}{\tabincell{c}{Fault\\level}}
&\multicolumn{5}{c|}{ADMM-HT v.s. ADMM-$l_1$}
&\multicolumn{5}{c|}{ADMM-MCP v.s. ADMM-$l_1$}\\
\cline{3-12}
& & \tabincell{c}{AVG \\diff.} & \tabincell{c}{standard\\deviation}
& t-value & p-value &\tabincell{c}{Confidence\\interval}
& \tabincell{c}{AVG \\diff.} & \tabincell{c}{standard\\deviation}
& t-value & p-value &\tabincell{c}{Confidence\\interval} \\
\hline
\multirow{3}{*}{ABA} &$0.005$ &0.244 &0.073 &15.0 &2.75E-12 &[0.210,0.278] &0.199 &0.065 &13.7 &1.33E-11
 &[0.169,0.23]\\ \cline{2-12}
&$ 0.01$ &0.302 &0.092 &14.8 &3.66E-12 &[0.256,0.349] &0.267 &0.092 &12.9 &3.55E-11
 &[0.22,0.313]\\ \cline{2-12}
&$ 0.05$ &0.634 &0.112 &25.3 &2.14E-16 &[0.578,0.691] &0.620 &0.111 &25.0 &2.68E-16 &[0.564,0.676]\\ \hline
\multirow{3}{*}{ASN} &$ 0.005$ &0.00127 &0.00030 &19.3 &2.93E-14 &[0.0011,0.0014] &0.00105 &0.00033 &14.3 &6.51E-12 &[0.0009,0.0012]\\ \cline{2-12}
&$ 0.01$ &0.00114 &0.00032 &16.1 &8.05E-13 &[0.0010,0.0013] &0.00115 &0.00036 &14.5 &4.88E-12 &[0.0010,0.0013]\\ \cline{2-12}
&$ 0.05$ &0.00094 &0.00046 &9.0 &1.30E-08 &[0.0007,0.0012] &0.00080 &0.00042 &8.5 &3.29E-08 &[0.0006,0.0010]\\ \hline
\multirow{3}{*}{\tabincell{c}{HOUS \\-ING}} &$ 0.005$ &0.0007 &0.00044 &7.3 &2.99E-07 &[0.00051,0.00092] &0.00056 &0.00044 &5.7 &8.50E-06 &[0.00035,0.00076]\\ \cline{2-12}
&$ 0.01$ &0.0010 &0.00053 &8.5 &3.34E-08 &[0.00075,0.00125] &0.00086 &0.00050 &7.7 &1.59E-07 &[0.00063,0.00110]\\ \cline{2-12}
&$ 0.05$ &0.0021 &0.00085 &11.3 &3.76E-10 &[0.00174,0.00253] &0.00208 &0.00085 &11.0 &5.39E-10 &[0.00169,0.00248]\\ \hline
\multirow{3}{*}{CON} &$ 0.005$ &0.00120 &0.00042 &12.7 &4.84E-11 &[0.00100,0.00140] &0.00116 &0.00041 &12.8 &4.50E-11 &[0.00097,0.00136]\\ \cline{2-12}
&$ 0.01$ &0.00137 &0.00040 &15.3 &1.89E-12 &[0.00118,0.00156] &0.00102 &0.00041 &11.0 &5.21E-10 &[0.00082,0.00121]\\ \cline{2-12}
&$ 0.05$ &0.00154 &0.00063 &11.0 &5.40E-10 &[0.00125,0.00184] &0.00186 &0.00062 &13.3 &2.23E-11 &[0.00156,0.00215]\\ \hline
\multirow{3}{*}{\tabincell{c}{ENER \\-GY}} &$ 0.005$ &0.00040 &0.00023 &7.8 &1.2E-07 &[0.0003,0.000514] &0.00042 &0.00023 &8.2 &6.01E-08 &[0.00031,0.00053]\\ \cline{2-12}
&$ 0.01$ &0.00047 &0.00025 &8.3 &4.8E-08 &[0.00035,0.00059] &0.00050 &0.00026 &8.7 &2.39E-08 &[0.00038,0.00062]\\ \cline{2-12}
&$ 0.05$ &0.00095 &0.00041 &10.3 &1.6E-09 &[0.00076,0.00114] &0.00095 &0.00042 &10.1 &2.12E-09 &[0.00076,0.00115]\\ \hline
\multirow{3}{*}{WQW} &$ 0.005$ &0.00035 &0.000128 &12.1 &1.06E-10 &[0.00029,0.00041] &0.00037 &0.000130 &12.5 &6.30E-11 &[0.00031,0.00043]\\ \cline{2-12}
&$ 0.01$ &0.00036 &0.000127 &12.6 &5.76E-11 &[0.0003,0.000415] &0.00041 &0.000151 &12.1 &1.21E-10 &[0.00034,0.00048]\\ \cline{2-12}
&$ 0.05$ &0.00033 &8.66E-05 &17.0 &2.78E-13 &[0.00029,0.00037] &0.00034 &0.000107 &14.0 &9.45E-12 &[0.00029,0.00039]\\ \hline
\end{tabular}
\caption{Results of paired t-test between the proposed algorithms and ADMM-$l_1$.}
\label{t-test1}
\end{table*}

\begin{table*}[tbp]
\newcommand{\tabincell}[2]{\begin{tabular}{@{}#1@{}}#2\end{tabular}}
\centering
\renewcommand{\multirowsetup}{\centering}
\begin{tabular}{|c|c|c|c|c|c|c|c|c|c|c|c|}
\hline
\multirow{2}{*}{Dataset} &\multirow{2}{*}{\tabincell{c}{Fault\\level}}
&\multicolumn{5}{c|}{ADMM-HT v.s. OLS}
&\multicolumn{5}{c|}{ADMM-MCP v.s. OLS}\\
\cline{3-12}
& & \tabincell{c}{AVG \\diff.} & \tabincell{c}{standard\\deviation}
& t-value & p-value &\tabincell{c}{Confidence\\interval}
& \tabincell{c}{AVG \\diff.} & \tabincell{c}{standard\\deviation}
& t-value & p-value &\tabincell{c}{Confidence\\interval} \\
\hline
\multirow{3}{*}{ABA} &$0.005$ &0.210 &0.084 &11.2 &4.05E-10 &[0.17,0.249] &0.165 &0.0887 &8.3 &4.54E-08 &[0.124,0.207] \\ \cline{2-12}
&$ 0.01$ &0.219 &0.065 &15.0 &2.64E-12 &[0.186,0.251] &0.183 &0.0699 &11.7 &1.99E-10 &[0.148,0.218]\\ \cline{2-12}
&$ 0.05$ &0.267 &0.048 &24.8 &3.04E-16 &[0.242,0.291] &0.253 &0.0487 &23.2 &1.04E-15 &[0.228,0.277]\\ \hline
\multirow{3}{*}{ASN} &$0.005$ &0.00319 &0.00099 &14.4 &5.50E-12 &[0.00273,0.00366] &0.00298 &0.00104 &12.8 &4.44E-11 &[0.00249,0.00347]\\ \cline{2-12}
&$ 0.01$ &0.00331 &0.00081 &18.3 &8.30E-14 &[0.00293,0.00368] &0.00368 &0.00078 &19.0 &3.93E-14 &[0.00295,0.00368]\\ \cline{2-12}
&$ 0.05$ &0.00270 &0.00069 &17.6 &1.66E-13 &[0.00237,0.00302] &0.00256 &0.00068 &16.7 &3.97E-13 &[0.00224,0.00288]\\ \hline
\multirow{3}{*}{\tabincell{c}{HOUS \\-ING}} &$0.005$ &0.00143 &0.00062 &10.4 &1.37E-09 &[0.00114,0.00172] &0.00127 &0.00073 &7.9 &1.12E-07 &[0.00093,0.00161]\\ \cline{2-12}
&$ 0.01$ &0.00190 &0.00071 &11.9 &1.46E-10 &[0.00157,0.00223] &0.00176 &0.00072 &10.9 &6.54E-10 &[0.00142,0.00210]\\ \cline{2-12}
&$ 0.05$ &0.00347 &0.00109 &14.2 &7.29E-12 &[0.00296,0.00399] &0.00340 &0.00107 &14.2 &6.82E-12 &[0.00290,0.00392]\\ \hline
\multirow{3}{*}{CON} &$ 0.005$ &0.0054 &0.00429 &5.7 &9.76E-06 &[0.0034,0.0074] &0.00538 &0.00430 &5.6 &1.17E-05 &[0.0034,0.0074]\\ \cline{2-12}
&$ 0.01$ &0.0063 &0.00488 &5.7 &7.89E-06 &[0.0040,0.0085] &0.00590 &0.00485 &5.4 &1.49E-05 &[0.0036,0.0082]\\ \cline{2-12}
&$ 0.05$ &0.0076 &0.00615 &5.5 &1.25E-05 &[0.0047,0.0105] &0.00790 &0.00610 &5.8 &6.70E-06 &[0.0051,0.0108] \\ \hline
\multirow{3}{*}{\tabincell{c}{ENER \\-GY}} &$ 0.005$ &0.00049 &0.000229 &9.6 &5.44E-09 &[0.00038,0.00060] &0.00050 &0.000227 &9.9 &3.14E-09 &[0.0004,0.0006]\\ \cline{2-12}
&$ 0.01$ &0.00069 &0.000241 &12.9 &3.80E-11 &[0.00058,0.00081] &0.00072 &0.000260 &12.3 &8.96E-11 &[0.0006,0.0009]\\ \cline{2-12}
&$ 0.05$ &0.00088 &0.000239 &16.4 &5.40E-13 &[0.00077,0.00099] &0.00088 &0.000249 &15.8 &1.06E-12 &[0.0008,0.0010]\\ \hline
\multirow{3}{*}{WQW} &$ 0.005$ &0.00089 &0.000290 &13.6 &1.52E-11 &[0.00075,0.00102] &0.00090 &0.000286 &14.2 &7.35E-12 &[0.00077,0.00104]\\ \cline{2-12}
&$ 0.01$ &0.00088 &0.000276 &14.3 &6.29E-12 &[0.00075,0.00101] &0.00093 &0.000269 &15.5 &1.53E-12 &[0.00081,0.00106]\\ \cline{2-12}
&$ 0.05$ &0.00077 &0.000167 &20.6 &9.04E-15 &[0.00069,0.00085] &0.00069 &0.000103 &30.1 &8.50E-18 &[0.00064,0.00074]\\ \hline
\end{tabular}
\caption{Results of paired t-test between the proposed algorithms and OLS.}
\label{t-test2}
\end{table*}

Finally, we use the paired t-test to illustrate that compared with existing algorithms the improvement of our proposed algorithms are very significant.
From Fig.~\ref{fig:fault} and TABLE~\ref{resultFault}, the performance of SVR, $l_1$-reg., HOM and OFR are worse than the ADMM-$l_1$ and OLS method.
Hence, we only conduct the paired t-test between the proposed algorithms and ADMM-$l_1$ and OLS.
The results of the t-test are shown in TABLE~\ref{t-test1} and~\ref{t-test2}.
For the one-tailed test with $95\%$ level of confidence and 20 trials, the critical t-value is $1.729$.
We can see that all the test t-values are greater than $1.729$ and all p-values are smaller than $0.05$.
In other words, we have enough confidence to say that on average the proposed methods are better than the ADMM-$l_1$ and the OLS algorithm.
Besides, the confidence intervals in TABLE~\ref{t-test1} and TABLE~\ref{t-test2} do not include zero.
Therefore, we can have enough confidence to say that the improvements of our proposed algorithms are very significant.
\section{Conclusion} \label{section6}
In the paper, the fault tolerant RBF neural network training and its center selection problem are studied.
Based on the ADMM framework, two novel algorithms are proposed.
They are respectively ADMM-MCP and ADMM-HT.
Both of them can handle the two tasks simultaneously.
In the first method, the MCP function is introduced to select centers.
While, in the second method, an $l_0$-norm term is directly used for center selection, and the hard threshold operation is utilized to handle the $l_0$-norm term.
Both two methods can globally converge to a unique limit point under several mild conditions.
From the experimental results, the performance of our proposed approaches are superior to many state-of-the-art methods. The performance of our two algorithms are similar with each other. But ADMM-HT can directly select the number of centers without tuning any regularization parameter.

\appendices

%\section*{Acknowledgment}
\ifCLASSOPTIONcaptionsoff
  \newpage
\fi



%\bibliographystyle{IEEEtran}
%\begin{thebibliography}{1}
%  0.5em minus 0.4em\relax Harlow, England: Addison-Wesley, 1999.%
%\end{thebibliography}
\bibliographystyle{IEEEtran}
\bibliography{IEEEabrv,my_reference}

\end{document}